\documentclass[10pt,twocolumn,letterpaper]{article}

\usepackage{titling}
\usepackage{cvpr}
\usepackage{times}
\usepackage{epsfig}
\usepackage{graphicx}
\usepackage{amsmath}
\usepackage{amssymb}
\usepackage{textcomp}
\usepackage{array}
\usepackage{multirow}
\usepackage{booktabs}
\usepackage[flushleft]{threeparttable}
\usepackage{tabularx}
\usepackage{subcaption}
\usepackage{rotating}
\usepackage[dvipsnames]{xcolor}

\usepackage{bm}
\usepackage{nicefrac}

\newcommand{\tabr}[2] {\multirow{#1}{*}{\rotatebox{90}{#2}}}

\newcommand{\vhad}{PHAV}

\usepackage[pagebackref=true,breaklinks=true,letterpaper=true,colorlinks,bookmarks=false]{hyperref}

\cvprfinalcopy 

\def\cvprPaperID{1978} 

\begin{document}


\title{Procedural Generation of Videos to Train Deep Action Recognition Networks}

\author{C\'{e}sar Roberto de Souza\textsuperscript{1,2}, Adrien Gaidon\textsuperscript{3}, Yohann Cabon\textsuperscript{1}, Antonio Manuel L\'{o}pez\textsuperscript{2} \\
	\textsuperscript{1}Computer Vision Group, NAVER LABS Europe, Meylan, France \\
	\textsuperscript{2}Centre de Visi\'{o} per Computador, Universitat Aut\`{o}noma de Barcelona, Bellaterra, Spain \\
	\textsuperscript{3}Toyota Research Institute, Los Altos, CA, USA \\
	{\tt\small \{cesar.desouza, yohann.cabon\}@europe.naverlabs.com, 
		adrien.gaidon@tri.global, antonio@cvc.uab.es}
}

\maketitle

\def\eg{\textit{e.g.,~}}
\def\cf{\textit{cf.~}}
\def\vs{\textit{vs.~}}
\def\ie{\textit{i.e.~}}
\def\wrt{\textit{w.r.t.~}}
\def\etal{\textit{et~al.~}}
\def\iid{\textit{i.i.d.~}}

\def\g{\gamma}
\def\d{\delta}
\def\p{\phi}
\def\s{\sigma}
\def\Re{\mathbb R}

\newcommand{\mocap}{MOCAP}

\begin{abstract}
Deep learning for human action recognition in videos is making significant
progress, but is slowed down by its dependency on expensive manual labeling of
large video collections.
In this work, we investigate the generation of synthetic training data for
action recognition, as it has recently shown promising results for a variety of other
computer vision tasks.
We propose an interpretable parametric generative model of human action videos
that relies on procedural generation and other computer graphics techniques of
modern game engines.
We generate a diverse, realistic, and physically plausible dataset of human
action videos, called PHAV for "Procedural Human Action Videos". It contains a
total of $39,982$ videos, with more than $1,000$ examples for each action of $35$
categories.  Our approach is not limited to existing motion capture sequences,
and we procedurally define $14$ synthetic actions.
We introduce a deep multi-task representation learning architecture to mix
synthetic and real videos, even if the action categories differ.
Our experiments on the UCF101 and HMDB51 benchmarks suggest that combining our
large set of synthetic videos with small real-world datasets can boost
recognition performance, significantly outperforming fine-tuning
state-of-the-art unsupervised generative models of videos.
\end{abstract}

{
\let\thefootnote\relax\footnote{Work done while A. Gaidon was at Xerox Research Centre Europe. Xerox Research Centre Europe was acquired by NAVER LABS and became NAVER LABS Europe after the original publication.}
\setcounter{footnote}{0}
}


\section{Introduction}\label{sec:intro}

Understanding human behavior in videos is a key problem in computer vision. Accurate representations of both appearance and motion require either carefully handcrafting features with prior knowledge (\eg the dense trajectories of~\cite{Wang2013a}) or end-to-end deep learning of high capacity models with a large amount of labeled data (\eg the two-stream network of~\cite{Simonyan2014}). These two families of methods have complementary strengths and weaknesses, and they often need to be combined to achieve state-of-the-art action recognition performance~\cite{Wang2015d,DeSouza2016}.
Nevertheless, deep networks have the potential to significantly improve their accuracy based on training data.
Hence, they are becoming the de-facto standard for recognition problems where it is possible to collect large labeled training sets, often by crowd-sourcing manual annotations (\eg ImageNet~\cite{DengCVPR09Imagenet}, MS-COCO~\cite{LinECCV14Microsoft}).
However, manual labeling is costly, time-consuming, error-prone, raises privacy concerns, and requires massive human intervention for every new task. This is often impractical, especially for videos, or even unfeasible for ground truth modalities like optical flow or depth. 

\begin{figure}[t!]
    \begin{center}
        \includegraphics[width=1.0\linewidth]{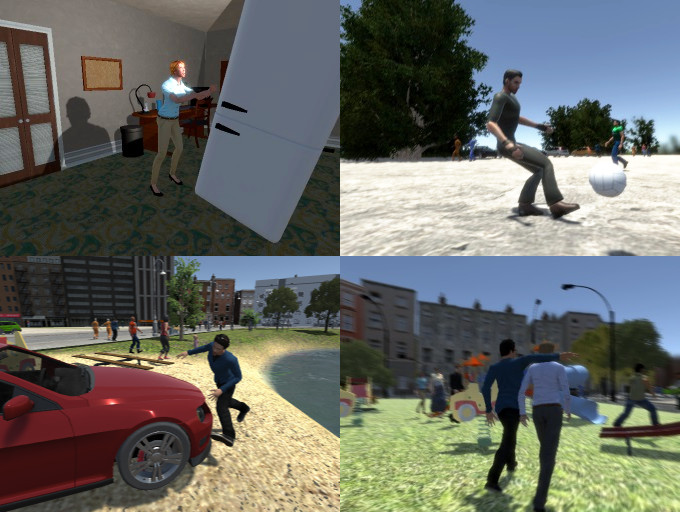}
        \caption{Procedurally generated human action videos (clockwise): push, kick ball, walking hug, car hit.}
        \label{fig:introframes}
    \end{center}
	\vspace{-8mm}
\end{figure}

Using realistic synthetic data generated from virtual worlds alleviates these issues. Thanks to modern modeling, rendering, and simulation software, virtual worlds allow for the efficient generation of vast amounts of controlled and algorithmically labeled data, including for modalities that cannot be labeled by a human. This approach has recently shown great promise for deep learning across a breadth of computer vision problems, including optical flow~\cite{MayerCVPR16Large}, depth estimation~\cite{LinECCV14Microsoft}, object detection~\cite{MarinCVPR10Learning, VazquezPAMI14Virtual, XuITS14Learning, SunBMVC14Virtual, PengICCV15Learning}, pose and viewpoint estimation~\cite{ShottonCVPR11Realtime, PaponICCV15Semantic, SuICCV16Render}, tracking~\cite{Gaidon2016}, and semantic segmentation~\cite{HandaCVPR16Understanding, RosCVPR16Synthia, RichterECCV16Playing}.

In this work, we investigate \emph{procedural generation of synthetic human action videos} from virtual worlds in order to train deep action recognition models. This is an open problem with formidable technical challenges, as it requires a full generative model of videos with realistic appearance and motion statistics conditioned on specific action categories. Our experiments suggest that our procedurally generated action videos can complement scarce real-world data. 
We report significant performance gains on target real-world categories although they differ from the actions present in our synthetic training videos.


Our first contribution is a \emph{parametric generative model of human action videos} relying on physics, scene composition rules, and procedural animation techniques like "ragdoll physics" that provide a much stronger prior than just viewing videos as tensors or sequences of frames. We show how to procedurally generate physically plausible variations of different types of action categories obtained by \mocap~datasets, animation blending, physics-based navigation, or entirely from scratch using programmatically defined behaviors. We use naturalistic actor-centric randomized camera paths to film the generated actions with care for physical interactions of the camera. Furthermore, our manually designed generative model has \emph{interpretable parameters} that allow to either randomly sample or precisely control discrete and continuous scene (weather, lighting, environment, time of day, etc), actor, and action variations to generate large amounts of diverse, physically plausible, and realistic human action videos.

Our second contribution is a quantitative experimental validation using a
modern and accessible game engine (Unity\textregistered Pro) to synthesize a
labeled dataset of $39,982$ videos, corresponding to more than
$1,000$ examples for each of $35$ action categories: $21$ grounded in
\mocap \ data, and $14$ entirely synthetic ones defined procedurally. Our
dataset, called \emph{PHAV} for "Procedural Human Action Videos" (\cf
Figure~\ref{fig:introframes} for example frames), is publicly available
for download\footnote{Data and tools available in \url{http://adas.cvc.uab.es/phav/}}. Our procedural generative model
took approximately $2$ months of $2$ engineers to be programmed and our PHAV
dataset $3$ days to be generated using $4$ gaming GPUs.
We investigate the use of this data in conjunction with the standard
UCF101~\cite{Soomro2012} and HMDB51~\cite{Kuehne2011} action recognition
benchmarks.
To allow for generic use, and as predefined procedural action categories may
differ from unknown a priori real-world target ones, we propose a multi-task
learning architecture based on the recent Temporal Segment Network~\cite{Wangb}
(TSN). We call our model \emph{Cool-TSN} (\cf Figure~\ref{fig:arch}) in
reference to the "cool world" of~\cite{VazquezNIPSDATA11Cool}, as we mix both
synthetic and real samples at the mini-batch level during training.
Our experiments show that the generation of our synthetic human action videos
can significantly improve action recognition accuracy, especially with small
real-world training sets, in spite of differences in appearance, motion, and
action categories.  Moreover, we outperform other state-of-the-art
generative video models~\cite{VondrickNIPS2016Generating} when combined with
the same number of real-world training examples.

The rest of the paper is organized as follows. Section~\ref{sec:related} presents a brief review of related works. In Section~\ref{sec:dataset}, we present our parametric generative model, and how we use it to procedurally generate our PHAV dataset. Section~\ref{sec:model} presents our cool-TSN deep learning algorithm for action recognition. We report our quantitative experiments in Section~\ref{sec:experiments} measuring the usefulness of PHAV. Finally, conclusions are drawn in Section~\ref{sec:conclusion}.


\section{Related work}\label{sec:related}

Synthetic data has been used to train visual models for object
detection and recognition, pose estimation, and indoor scene understanding
\cite{MarinCVPR10Learning, VazquezPAMI14Virtual,
XuITS14Learning,XuPAMI14Domain, ShottonCVPR11Realtime, PishchulinCVPR11Learning,
PepikCVPR12Teaching, SatkinBMVC12Data,
AubryCVPR14Seeing, ChenCVPR14Beat, PaponICCV15Semantic,
PengICCV15Learning, HandaX15SynthCam3D,
HattoriCVPR15Learning, MassaCVPR16Deep, SuICCV163D,
SuICCV16Render, HandaCVPR16Understanding}.
\cite{HaltakovGCPR13Framework} used a virtual racing circuit to generate
different types of pixel-wise ground truth (depth, optical flow and class
labels).
\cite{RosCVPR16Synthia, RichterECCV16Playing} relied on game technology to
train deep semantic segmentation networks, while \cite{Gaidon2016} used it for
multi-object tracking, \cite{ShafaeiBMVC16Play} for depth estimation from RGB,
and \cite{Sizikova1ECCV16Enhancing} for place recognition.
Several works use synthetic scenarios to evaluate the performance of different
feature descriptors \cite{KanevaICCV11Evaluating, AubryICCV15Understanding,
VeeravasarapuX15Simulations, VeeravasarapuX15Model, VeeravasarapuX16Model} and
to train and test optical and/or scene flow estimation methods
\cite{MeisterCEMT11Real, Butler2012, OnkarappaMTA15Synthetic,
MayerCVPR16Large}, stereo algorithms \cite{HaeuslerGCPR13Synthesizing}, or 
trackers~\cite{TaylorCVPR07OVVV, Gaidon2016}.
They have also been used for learning artificial behaviors
such as playing Atari games \cite{MnihNIPSWDL13Playing}, imitating 
players in shooter games \cite{LlarguesESA14Artificial}, end-to-end
driving/navigating \cite{ChenICCV15DeepDriving, ZhuX16Target}, learning common
sense \cite{VedantamICCV15Learning, ZitnickPAMI16Adopting} or physical
intuitions~\cite{LererICML16Learning}.
Finally, virtual worlds have also been explored from an animator's perspective. Works in computer graphics have investigated producing animations from sketches~\cite{Guay2015}, using physical-based models to add motion to sketch-based animations~\cite{Guay2015a}, and creating constrained camera-paths~\cite{Galvane2015}. 
However, due to the formidable complexity of realistic animation, video
generation, and scene understanding, these approaches focus on basic controlled
game environments, motions, and action spaces. 

To the best of our knowledge, ours is the first work to investigate virtual
worlds and game engines to generate synthetic training videos for action
recognition. Although some of the aforementioned related works rely on virtual
characters, their actions are not the focus, not procedurally generated, and
often reduced to just walking.

The related work of~\cite{MatikainenICCV11Feature} uses \mocap~data to induce
realistic motion in an "abstract armature" placed in an empty synthetic
environment, generating $2,000$ short 3-second clips at 320x240 and 30FPS. From
these non-photo-realistic clips, handcrafted motion features are selected as
relevant and later used to learn action recognition models for 11 actions in
real-world videos.  Our approach \emph{does not just replay \mocap, but
procedurally generates new action categories} -- including interactions between
persons, objects and the environment -- as well as random \emph{physically
plausible} variations.  Moreover, we jointly generate and learn deep
representations of both action appearance and motion thanks to our realistic
synthetic data, and our multi-task learning formulation to combine real and
synthetic data. 

A recent alternative to our procedural generative model that also does not
require manual video labeling is the unsupervised Video Generative Adversarial
Network (VGAN) of~\cite{VondrickNIPS2016Generating}. Instead of leveraging
prior structural knowledge about physics and human actions, the authors view
videos as tensors of pixel values and learn a two-stream GAN on $5,000$ hours
of unlabeled Flickr videos. This method focuses on tiny videos and capturing
scene motion assuming a stationary camera. This architecture can be used for
action recognition in videos when complemented with prediction layers
fine-tuned on labeled videos. Compared to this approach, our proposal allows to
work with any state-of-the-art discriminative architecture, as video generation
and action recognition are decoupled steps. We can, therefore, benefit from a
strong ImageNet initialization for both appearance and motion streams as
in~\cite{Wangb}. Moreover, we can decide what specific
actions/scenarios/camera-motions to generate, enforcing diversity thanks to our
interpretable parametrization. For these reasons, we show in
Section~\ref{sec:experiments} that, given the same amount of labeled videos,
our model achieves nearly two times the performance of the unsupervised 
features shown in~\cite{VondrickNIPS2016Generating}.


\section{PHAV: Procedural Human Action Videos}\label{sec:dataset}

In this section we introduce our interpretable parametric generative model of
videos depicting particular human actions, and how we use it to generate our
PHAV dataset. We describe the procedural generation techniques we
leverage to randomly sample diverse yet physically plausible appearance and
motion variations, both for \mocap-grounded actions and programmatically defined
categories.

\begin{figure}[t!]
	\begin{center}
		\includegraphics[width=\columnwidth]{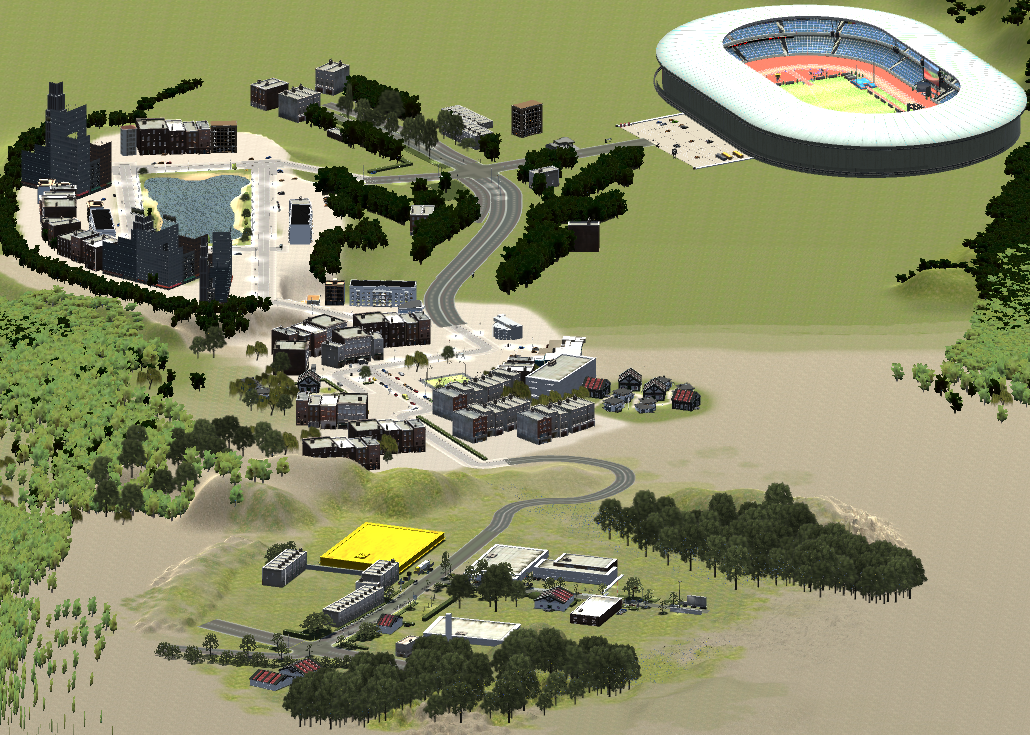}
		\caption{Orthographic view of different world regions.}
		\label{fig:world}
	\end{center}
	\vspace*{-5mm}
\end{figure}

\subsection{Action scene composition}

In order to generate a human action video, we place a \textit{protagonist} 
performing an \textit{action} in an \textit{environment}, under 
particular \textit{weather conditions} at a specific \textit{period}
of the day. There can be one or more \textit{background actors}
in the scene, as well as one or more \textit{supporting characters}. We film 
the virtual scene using a parametric \textit{camera behavior}.

The protagonist is the main human model performing the action. For actions
involving two or more people, one is chosen to be the protagonist.
Background actors can freely walk in the current virtual environment, while
supporting characters are actors with a secondary role necessary to complete an
action, \eg hold hands.

The action is a human motion belonging to a predefined semantic category
originated from one or more motion data sources (described in
section~\ref{s:actions}), including pre-determined motions from a
\mocap~dataset, or programmatic actions defined using procedural animation
techniques~\cite{Egges2008, VanWelbergen2009}, in particular ragdoll physics.
In addition, we use these techniques to sample physically-plausible motion
variations (described in section~\ref{s:movars}) to increase diversity.

The environment refers to a region in the virtual world (\cf
Figure~\ref{fig:world}), which consists of large urban areas, natural
environments (\eg forests, lakes, and parks), indoor scenes, and sports grounds
(\eg a stadium). Each of these environments may contain moving or static
background pedestrians or objects -- \eg cars, chairs -- with which humans can
physically interact, voluntarily or not.
The outdoor weather in the virtual world can be rainy, overcast, clear, or
foggy.
The period of the day can be dawn, day, dusk, or night.

Similar to~\cite{Gaidon2016,RosCVPR16Synthia}, we use a library of pre-made 3D models obtained
from the Unity Asset Store, which includes artist-designed human, object, and
texture models, as well as semi-automatically created realistic environments
(\eg selected scenes from the VKITTI dataset~\cite{Gaidon2016}).

\subsection{Camera}

\begin{figure}[t!]
    \begin{center}
        \begin{subfigure}{0.55\columnwidth}
            \includegraphics[width=\textwidth,trim={1cm 0 1cm 1.5cm}]{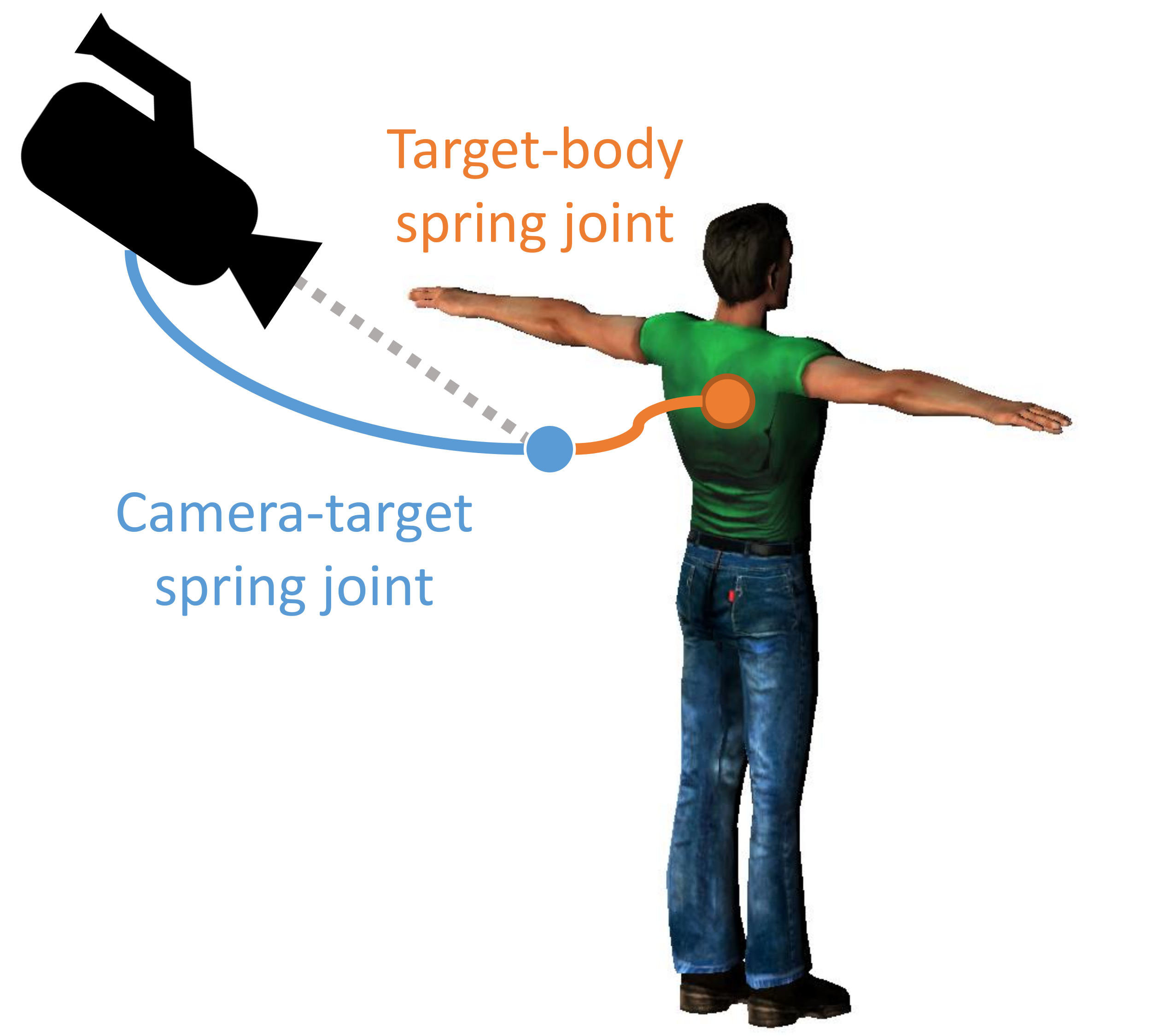}
        \end{subfigure}
        \begin{subfigure}{0.40\columnwidth}
            \includegraphics[width=\textwidth]{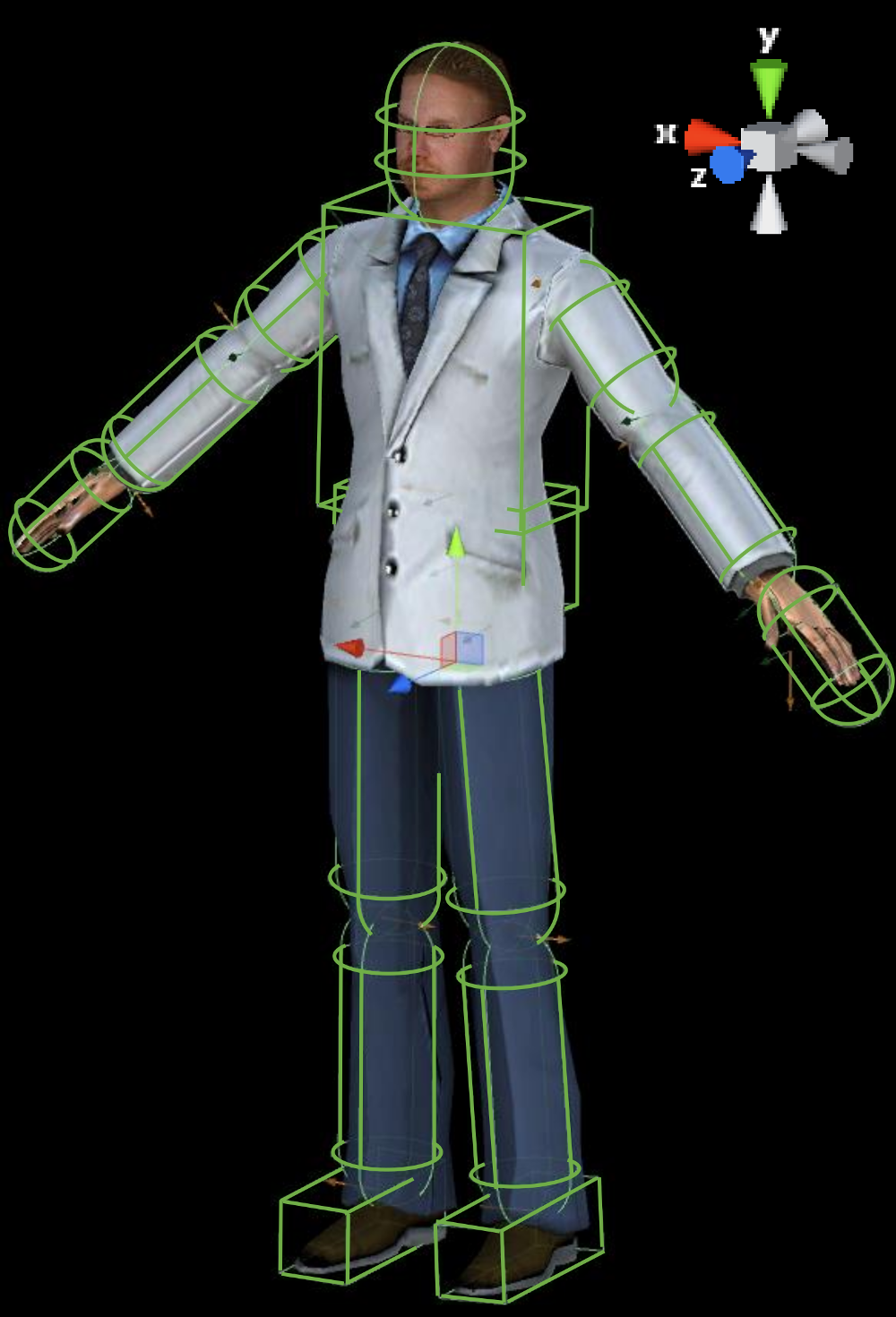}
        \end{subfigure}
        \caption{Left: schematic representation of our Kite camera.
            Right: human ragdoll configuration with 15 muscles.}
        \label{fig:kite}
    \end{center}
    \vspace*{-6mm}
\end{figure}

We use a physics-based camera which we call the Kite camera (\cf
Figure~\ref{fig:kite}) to track the protagonist in a scene. This physics-aware
camera is governed by a rigid body attached by a spring to a target position
that is, in turn, attached to the protagonist by another spring. By randomly
sampling different parameters for the drag and weight of the rigid bodies, as
well as elasticity and length of the springs, we can achieve cameras with a
wide range of shot types, 3D transformations, and tracking behaviors, such as
following the actor, following the actor with a delay, or stationary. Another
parameter controls the direction and strength of an initial impulse that starts
moving the camera in a random direction. With different rigid body parameters,
this impulse can cause our camera to simulate a handheld camera, move in a
circular trajectory, or freely bounce around in the scene while filming the
attached protagonist.

\subsection{Actions}\label{s:actions}

Our approach relies on two main existing data sources for basic human
animations.
First, we use the CMU \mocap~database \cite{CarnegieMellonGraphicsLab2016},
which contains 2605 sequences of 144 subjects divided in 6 broad categories, 23
subcategories and further described with a short text. We leverage relevant motions
from this dataset to be used as a motion source for our procedural generation
based on a simple filtering of their textual motion descriptions.
Second, we use a large amount of hand-designed realistic motions made by
animation artists and available on the Unity Asset Store.

The key insight of our approach is that \emph{these sources need not
necessarily contain motions from predetermined action categories of interest,
neither synthetic nor target real-world actions (unknown a priori)}.
Instead, we propose to use these sources to form a \emph{library of atomic motions} to procedurally generate realistic action categories.
We consider atomic motions as individual movements of a limb in a larger animation
sequence. For example, atomic motions in a "walk" animation include movements such as rising a left leg, rising a right leg, and pendular arm movements. 
Creating a library of atomic motions enables us to later recombine those atomic actions into new higher-level animation sequences, e.g., "hop" or "stagger".

Our PHAV dataset contains 35 different action classes (\cf
Table~\ref{tab:categories}), including 21 simple categories present in
HMDB51 and composed directly of some of the aforementioned atomic
motions. In addition to these actions, we programmatically define 10 action
classes involving a single actor and 4 action classes involving two person
interactions.
We create these new synthetic actions by taking atomic motions as a base and
using procedural animation techniques like blending and ragdoll physics (\cf
Section~\ref{s:movars}) to compose them in a physically plausible manner
according to simple rules defining each action, such as tying hands together (\eg
"walk hold hands"), disabling one or more muscles (\eg "crawl", "limp"), or
colliding the protagonist against obstacles (\eg "car hit", "bump into each other").

\begin{table}[t!]
\small
    \begin{center}
        \begin{tabular}{>{\centering\bfseries}m{5.8em} >{\centering}m{0.8em} >{\centering\arraybackslash}m{14.9em}}
            \toprule
            Type & \# & Actions \\
            \midrule
            sub-HMDB & 21 & brush hair, catch, clap, climb stairs, golf, jump, kick ball, push, pick, pour, pull up, run, shoot ball, shoot bow, shoot gun, sit, stand, swing baseball, throw, walk, wave \\
            One-person synthetic & 10 & car hit, crawl, dive floor, flee, hop, leg split, limp, moonwalk, stagger, surrender \\
            Two-people synthetic & 4 & walking hug, walk hold hands, walk the line, bump into each other \\
            \bottomrule
        \end{tabular}
    \end{center}
    \vspace*{-6mm}
    \caption{Actions included in our PHAV dataset. }
    \label{tab:categories}
    \vspace*{-6mm}
\end{table}

\subsection{Physically plausible motion variations}\label{s:movars}

We now describe procedural animation techniques~\cite{Egges2008,
VanWelbergen2009} to randomly generate large amounts of physically plausible
and diverse action videos, far beyond from what can be achieved by simply replaying source atomic motions.

\noindent \textbf{Ragdoll physics.} 
A key component of our work is the use of ragdoll physics. 
Ragdoll physics are limited real-time physical simulations 
that can be used to animate a model (such as a human model) 
while respecting basic physics properties such as connected
joint limits, angular limits, weight and strength.
We consider ragdolls with 15 movable body parts (referenced here as muscles),
as illustrated in Figure~\ref{fig:kite}.
For each action, we separate those
15 muscles into two disjoint groups: those that are strictly necessary for
performing the action, and those that are complementary (altering their
movement should not interfere with the semantics of the currently considered
action). The presence of the ragdoll allows us to introduce variations of
different nature in the generated samples. The other modes of variability
generation described in this section will assume that the physical plausibility
of the models is being kept by the use of ragdoll physics. We use RootMotion's PuppetMaster\footnote{\url{http://root-motion.com}} for implementing and controlling human ragdolls in Unity\textregistered Pro.

\noindent \textbf{Random perturbations.} 
Inspired by \cite{Perlin1995}, we create variations of a given motion
by adding random perturbations to muscles that should not alter the 
semantic category of the action being performed. Those perturbations
are implemented by adding a rigid body to a random subset of the
complementary muscles. Those bodies are set to orbit around the muscle's
position in the original animation skeleton, drifting the movement of the
puppet's muscle to its own position in a periodic oscillating movement.
More detailed references on how to implement variations of this type can
be found in ~\cite{Perlin1995,Egges2008,Perlin2008,VanWelbergen2009} and
references therein.

\noindent \textbf{Muscle weakening.} 
We vary the strength of the avatar performing the action. By reducing its
strength, the actor performs an action with seemingly more difficulty.

\noindent \textbf{Action blending.} 
Similarly to modern video games, we use a blended ragdoll technique
to constrain the output of a pre-made animation to physically-plausible motions.
In action blending, we randomly sample a different motion sequence (coming
either from the same or from a different class) and replace the movements of
current complementary muscles with those from this new action.
We limit the number of blended actions in \vhad~to be at most two.

\begin{figure}[t!]
	\begin{center}
		\begin{subfigure}{0.15\textwidth}
			\includegraphics[width=\textwidth]{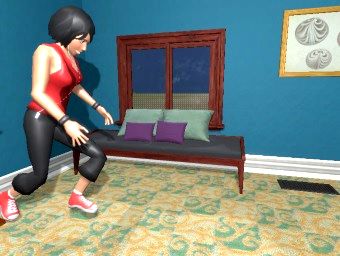}
		\end{subfigure}
		\begin{subfigure}{0.15\textwidth}
			\includegraphics[width=\textwidth]{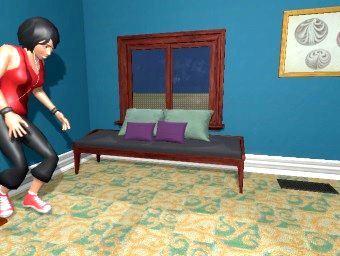}
		\end{subfigure}
		\begin{subfigure}{0.15\textwidth}
			\includegraphics[width=\textwidth]{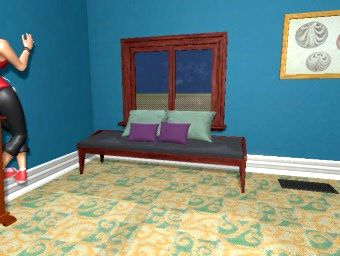}
		\end{subfigure}

		\begin{subfigure}{0.15\textwidth}
			\includegraphics[width=\textwidth]{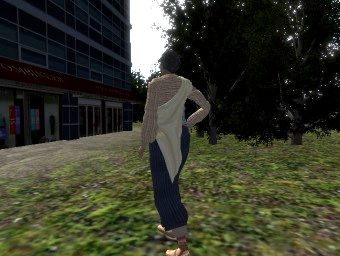}
		\end{subfigure}
		\begin{subfigure}{0.15\textwidth}
			\includegraphics[width=\textwidth]{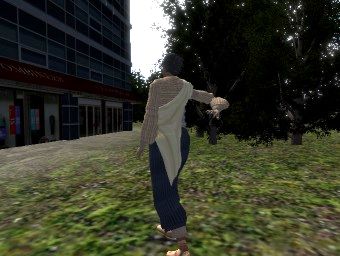}
		\end{subfigure}
		\begin{subfigure}{0.15\textwidth}
			\includegraphics[width=\textwidth]{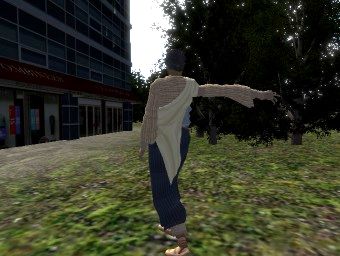}
		\end{subfigure}

		\vspace*{-2mm}
		\caption{
			Example generation failure cases. 
			First row: physics violations (passing through a wall).
			Second row: over-constrained joints and unintended variations.}
		\label{fig:failures}
		\vspace*{-7mm}
	\end{center}
\end{figure}

\noindent \textbf{Objects.} 
The last physics-based source of variation is the use of objects. First, we
manually annotated a subset of the \mocap~actions marking the instants in time
where the actor started or ended the manipulation of an object. Second, we use
inverse kinematics to generate plausible programmatic interactions.
For reproducibility, our annotated subset of MOCAP actions, as well as the code
for interacting with objects for particular actions will be available upon
request.

\noindent \textbf{Failure cases.} 
Although our approach uses physics-based procedural animation
techniques, unsupervised generation of large amounts of random variations with
a focus on diversity inevitably causes edge cases where physical models fail.
This results in glitches reminiscent of typical video game bugs (\cf
Figure~\ref{fig:failures}). Using a random $1\%$ sample of our dataset, we
manually estimated that this corresponds to less than $10\%$ of the videos 
generated.  Although this could be improved, our experiments in
Section~\ref{sec:experiments} show that this noise does not prevent us from
improving the training of deep action recognition networks using this data.

\noindent \textbf{Extension to complex activities.} 
Using ragdoll physics and a large enough library of atomic actions, it is
possible to create complex actions by hierarchical composition. For instance, our "Car Hit" action is procedurally defined
by composing atomic actions of a person (walking and/or doing other activities)
with those of a car (entering in a collision with the person), followed by the
person falling in a physically plausible fashion.
However, while atomic actions have been validated as an
effective decomposition for the recognition of potentially complex
actions~\cite{Gaidon2013a},
we have not studied how this approach would scale with the complexity
of the actions, notably due to the combinatorial nature of complex
events. We leave this as future work.

\subsection{Interpretable parametric generative model}

\begin{figure}[t!]
    \begin{center}
        \includegraphics[width=0.6\columnwidth]{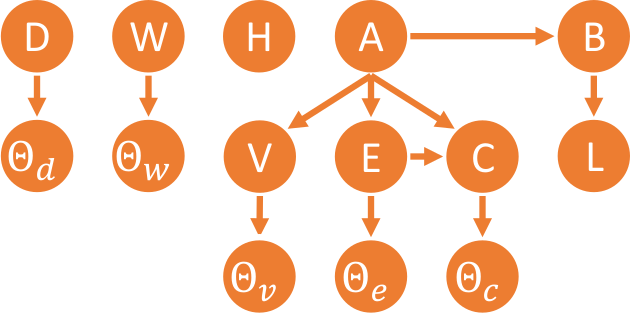}
        \caption{Simplified graphical model for our generator.}
        \label{fig:pgm}
    \end{center}
    \vspace*{-6mm}
\end{figure}

We define a human action video as a random variable
%
$X = \left \langle H, A, L, B, V, C, E, D, W \right \rangle$,
%
where
$H$ is a human model,
$A$ an action category,
$L$ a video length,
$B$ a set of basic motions (from \mocap, manual design, or programmed),
$V$ a set of motion variations,
$C$ a camera,
$E$ an environment,
$D$ a period of the day,
$W$ a weather condition, and
possible values for those parameters are shown in 
Table~\ref{tab:variations}. Our generative model (\cf~Figures \ref{fig:pgm},
\ref{fig:pgm1}, \ref{fig:pgm2}, and \ref{fig:pgm3})
for an action video $X$ is given by:
\begin{equation}
\begin{aligned}
P(X) = & P(H)~P(A)~P(L \mid B)~P(B \mid A)      \\         
       & P(\Theta_v \mid V)~P(V \mid A)       
         ~P(\Theta_e \mid E)~P(E \mid A)        \\
       & P(\Theta_c \mid C)~P(C \mid A, E)      \\
       & P(\Theta_d \mid D)~P(D)              
         ~P(\Theta_w \mid W)~P(W)      
\end{aligned}
\end{equation}

\begin{table}[t!]
    \vspace*{-2mm}
    \small
    \begin{center}
        \begin{tabular}{>{\bfseries}l >{\centering}m{0.8em} >{\arraybackslash}p{14em}}
            \toprule
            Parameter          & \# & Possible values \\
            \midrule
            Human Model (H)    & 20 & models designed by artists \\
            Environment (E)    &  7 & simple, urban, green, middle, lake, stadium, house interior \\
            Weather (W)        &  4 & clear, overcast, rain, fog \\
            Period of day (D)  &  4 & night, dawn, day, dusk \\
            Variation (V)      &  5 & $\varnothing$, muscle perturbation and weakening,
                                      action blending, objects \\ 
            \bottomrule
        \end{tabular}
    \end{center}
    \vspace*{-5mm}
    \caption{Overview of key random variables of our parametric generative
        model of human action videos (\cf section \ref{ss:supp-variables} of
        the attached supplementary material for more details).}
    \label{tab:variations}
\end{table}

\begin{figure*}[]
    \begin{center}
        \includegraphics[trim={0cm 0cm 0cm 0cm},clip,width=17.5cm]{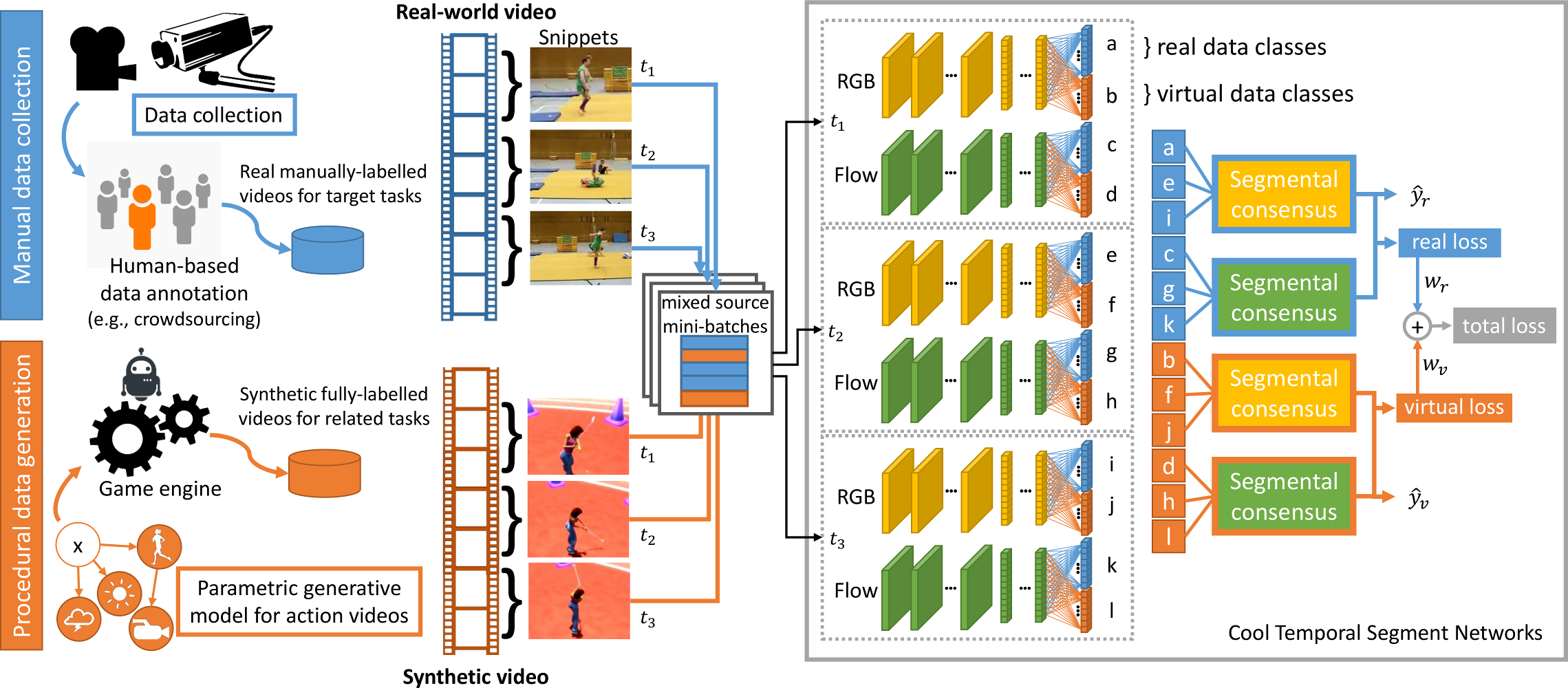}
        \caption{Our "Cool-TSN" deep multi-task learning architecture for end-to-end action recognition in videos.}
        \label{fig:arch}
    \end{center}
    \vspace{-5mm}
\end{figure*}

\noindent where $\Theta_w$ is a random variable (r.v.)\ on weather-specific parameters
(\eg intensity of rain, clouds, fog), $\Theta_c$ is a r.v.\ on camera-specific
parameters (\eg weights and stiffness for Kite camera springs), $\Theta_e$ is a
r.v.\ on environment-specific parameters (\eg current waypoint, waypoint
locations, background pedestrian starting points and destinations), $\Theta_d$
is a r.v.\ on period-specific parameters (\eg amount of sunlight, sun
orientation), and $\Theta_v$ is a r.v.\ on variation-specific parameters (\eg
strength of each muscle, strength of perturbations, blending muscles).
The probability functions associated with categorical variables (\eg $A$) can be either uniform, or configured manually to use pre-determined weights. 
Similarly, probability distributions associated with continuous values 
(\eg $\Theta_c$) are either set using a uniform distribution with finite support, or using triangular distributions with pre-determined support and most likely value. All values used are disclosed in the supplementary material.
We give additional details about our graphical model, as well as the values used to
configure the parameter distributions in section \ref{sec:supp-gen-model} of the
attached supplementary material.

\subsection{PHAV generation details}

We generate videos with lengths between 1 and 10 seconds, at 30FPS, and 
resolution of 340x256 pixels. 
We use anti-aliasing, motion blur, and standard photo-realistic cinematic
effects.
We have generated 55 hours of videos, with approximately 6M frames and
at least 1,000 videos per action category.
Our parametric model can generate fully-annotated action videos 
(including depth, flow, semantic segmentation, and human pose ground-truths) at 3.6
FPS using one consumer-grade gaming GPU (NVIDIA GTX 1070). In contrast, the average
annotation time for data-annotation methods such as
\cite{RichterECCV16Playing,CordtsCVPR16Cityscapes,BrostowPRL09Semantic} are
significantly below 0.5 FPS. While those works deal with semantic segmentation
where the cost of annotation is higher, we can generate all modalities for
roughly the same cost as RGB. We further note that all modalities are
included in our public dataset release.


\interfootnotelinepenalty=10000

\section{Cool Temporal Segment Networks}\label{sec:model}

We propose to demonstrate the usefulness of our PHAV dataset via deep
multi-task representation learning. 
Our main goal is to learn an end-to-end action recognition model for real-world
target categories by combining a few examples of labeled real-world videos with
a large number of procedurally generated videos for different surrogate
categories. Our hypothesis is that, although the synthetic examples differ in
statistics and tasks, their realism, quantity, and diversity can act as a
strong prior and regularizer against overfitting, towards data-efficient
representation learning that can operate with few manually labeled real videos.
Figure~\ref{fig:arch} depicts our learning algorithm inspired
by~\cite{Simonyan2014}, but adapted for the recent state-of-the-art Temporal
Segment Networks (TSN) of~\cite{Wangb} with "cool
worlds"~\cite{VazquezNIPSDATA11Cool}, \ie mixing real and virtual data during
training.

\subsection{Temporal Segment Networks} 

The recent TSN architecture of \cite{Wangb} improves significantly on the
original two-stream architecture of~\cite{Simonyan2014}. It processes both RGB
frames and stacked optical flow frames using a deeper Inception
architecture~\cite{Szegedy2014} with Batch Normalization~\cite{Ioffe2015} and
DropOut~\cite{Hinton2014}. Although it still requires massive labeled training
sets, this architecture is more data efficient, and therefore more suitable for
action recognition in videos.
In particular, \cite{Wangb} shows that both the appearance and motion streams of
TSNs can benefit from a strong initialization on ImageNet, which is one of
the main factors responsible for the high recognition accuracy of TSN.

Another improvement of TSN is the explicit use of long-range temporal structure
by jointly processing random short snippets from a uniform temporal subdivision
of a video.  TSN computes separate predictions for $K$ different temporal
segments of a video.  These partial predictions are then condensed into a
video-level decision using a segmental consensus function $\mathbf{G}$. We use
the same parameters as \cite{Wangb}: a number of segments $K=3$, and the
consensus function:
$\mathbf{G} = \frac{1}{K} \sum_{k=1}^K \mathcal{F}(T_k; W)$,
\noindent where $\mathcal{F}(T_k; W)$ is a function representing a CNN architecture with
weight parameters $W$ operating on short snippet $T_k$ from video segment $k$.

\subsection{Multi-task learning in a Cool World}\label{ss:multi_task}

As illustrated in Figure~\ref{fig:arch}, the main differences we introduce with
our "Cool-TSN" architecture are at both ends of the training procedure: (i) the
mini-batch generation, and (ii) the multi-task prediction and loss layers.

\noindent \textbf{Cool mixed-source mini-batches.} Inspired
by~\cite{VazquezNIPSDATA11Cool, RosCVPR16Synthia}, we build mini-batches
containing a mix of real-world videos and synthetic ones.
Following~\cite{Wangb}, we build minibatches of 256 videos divided in blocks of
32 dispatched across 8 GPUs for efficient parallel training using
MPI\footnote{\scriptsize\url{https://github.com/yjxiong/temporal-segment-networks}}. Each 32
block contains 10 random synthetic videos and 22 real videos in all our
experiments, as we observed it roughly balances the contribution of the
different losses during backpropagation.
Note that although we could use our generated ground truth flow for the PHAV
samples in the motion stream, we use the same fast optical flow estimation
algorithm as~\cite{Wangb} (TVL1~\cite{Zach2007ADB}) for all samples in order to
fairly estimate the usefulness of our generated videos.

\noindent \textbf{Multi-task prediction and loss layers.}
Starting from the last feature layer of each stream, we create two separate
computation paths, one for target classes from the real-world dataset, and
another for surrogate categories from the virtual world.
Each path consists of its own segmental consensus, fully-connected prediction,
and softmax loss layers.
As a result, we obtain the following multi-task loss:
\begin{equation}
\small
 \mathcal{L}(y, \mathbf{G}) = \sum_{z \in \{real, virtual\}} \delta_{\{y \in C_z\}} w_z \mathcal{L}_z(y, \mathbf{G}) 
\end{equation}
\vspace*{-2mm}
\begin{equation}
\small
 \mathcal{L}_z(y, \mathbf{G}) = - \sum_{i \in C_z} y_i \left( G_i - \log \sum_{j \in C_z} \exp{G_j} \right) 
\end{equation}
\noindent where $z$ indexes the source dataset (real or virtual) of the video,
$w_z$ is a loss weight (we use the relative proportion of $z$ in the mini-batch),
$C_z$ denotes the set of action categories for dataset $z$, and $\delta_{\{y
\in C_z\}}$ is the indicator function that returns one when label $y$ belongs
to $C_z$ and zero otherwise.
We use standard SGD with backpropagation to minimize that objective, and as
every mini-batch contains both real and virtual samples, every iteration is
guaranteed to update both shared feature layers and separate prediction layers
in a common descent direction.
We discuss the setting of the learning hyper-parameters (\eg learning rate,
iterations) in the following experimental section.


\section{Experiments}\label{sec:experiments}

In this section, we detail our action recognition experiments on widely used
real-world video benchmarks. We quantify the impact of multi-task
representation learning with our procedurally generated \vhad~videos on real-world
accuracy, in particular in the small labeled data regime. We also compare our
method with the state of the art on both fully supervised and unsupervised
methods.

\subsection{Real world action recognition datasets}

We consider the two most widely used real-world public benchmarks for human
action recognition in videos.
The \textbf{HMDB-51} \cite{Kuehne2011} dataset contains 6,849 fixed resolution 
videos clips divided between 51 action categories. The evaluation metric for
this dataset is the average accuracy over three data splits. 
The \textbf{UCF-101} \cite{Soomro2012,Jiang2013} dataset contains 13,320 video clips
divided among 101 action classes. Like HMDB, its standard evaluation metric is the
average mean accuracy over three data splits.
Similarly to UCF-101 and HMDB-51, we generate three random splits on our
\vhad~dataset, with 80\% for training and the rest for testing, and report average
accuracy when evaluating on \vhad.

\subsection{Temporal Segment Networks}

In our first experiments (\cf Table \ref{results_multi}), we reproduce
the performance of the original TSN in UCF-101 and HMDB-51 using
the same learning parameters as in~\cite{Wangb}. 
For simplicity, we use neither cross-modality pre-training nor a third
warped optical flow stream like~\cite{Wangb}, as their impact on TSN is limited
with respect to the substantial increase in training time and computational
complexity, degrading only by $-1.9\%$ on HMDB-51, and $-0.4\%$ on UCF-101.

We also estimate performance on \vhad~separately, and fine-tune \vhad~networks
on target datasets. Training and testing on \vhad~yields an average accuracy of 82.3\%, which is between that of HMDB-51 and UCF-101. 
This sanity check confirms that, just like real-world videos, our synthetic videos contain both appearance and motion patterns that can be captured by TSN to discriminate between our different procedural categories.
We use this network to perform fine-tuning experiments (TSN-FT), using
its weights as a starting point for training TSN on UCF101 and HMDB51 instead
of initializing directly from ImageNet as in~\cite{Wangb}. We discuss learning parameters and results below.

\begin{table}[]
    \small
    \centering
    \begin{tabular}{@{}ccccc@{}}
        \toprule
        Target  & Model        & Spatial & Temporal & Full          \\
\midrule
        \vhad   & TSN          & 65.9    & 81.5     & 82.3          \\
\midrule
        UCF-101 & \cite{Wangb} & 85.1    & 89.7     & 94.0          \\
        UCF-101 & TSN          & 84.2    & 89.3     & 93.6          \\
        UCF-101 & TSN-FT       & 86.1    & 89.7     & 94.1          \\
        UCF-101 & Cool-TSN     & 86.3    & 89.9     & \textbf{94.2} \\
\midrule
        HMDB-51 & \cite{Wangb} & 51.0    & 64.2     & 68.5          \\ 
        HMDB-51 & TSN          & 50.4    & 61.2     & 66.6          \\
        HMDB-51 & TSN-FT       & 51.0    & 63.0     & 68.9          \\
        HMDB-51 & Cool-TSN     & 53.0    & 63.9     & \textbf{69.5} \\
\bottomrule 
    \end{tabular}
    \vspace*{-2mm}
    \caption{Impact of our PHAV dataset using Cool-TSN. \cite{Wangb} uses TSN with cross-modality training.}
    \label{results_multi}
    \vspace*{-4mm}
\end{table}

\subsection{Cool Temporal Segment Networks}

In Table \ref{results_multi} we also report results of our Cool-TSN multi-task
representation learning, (Section~\ref{ss:multi_task}) which additionally uses
\vhad~to train UCF-101 and HMDB-51 models.
We stop training after $3,000$ iterations for RGB streams and $20,000$ for flow streams, all other parameters as in~\cite{Wangb}. 
Our results suggest that leveraging \vhad~through either Cool-TSN or TSN-FT yields recognition improvements for all modalities in all datasets, with advantages in using Cool-TSN especially for the smaller HMDB-51.
This provides quantitative experimental evidence supporting our claim that procedural generation of synthetic human action videos can indeed act as a strong prior (TSN-FT) and regularizer (Cool-TSN) when learning deep action recognition networks.

\begin{figure}[]
    \begin{center}
\vspace*{-5mm}
        \includegraphics[width=1.01\columnwidth
        ]{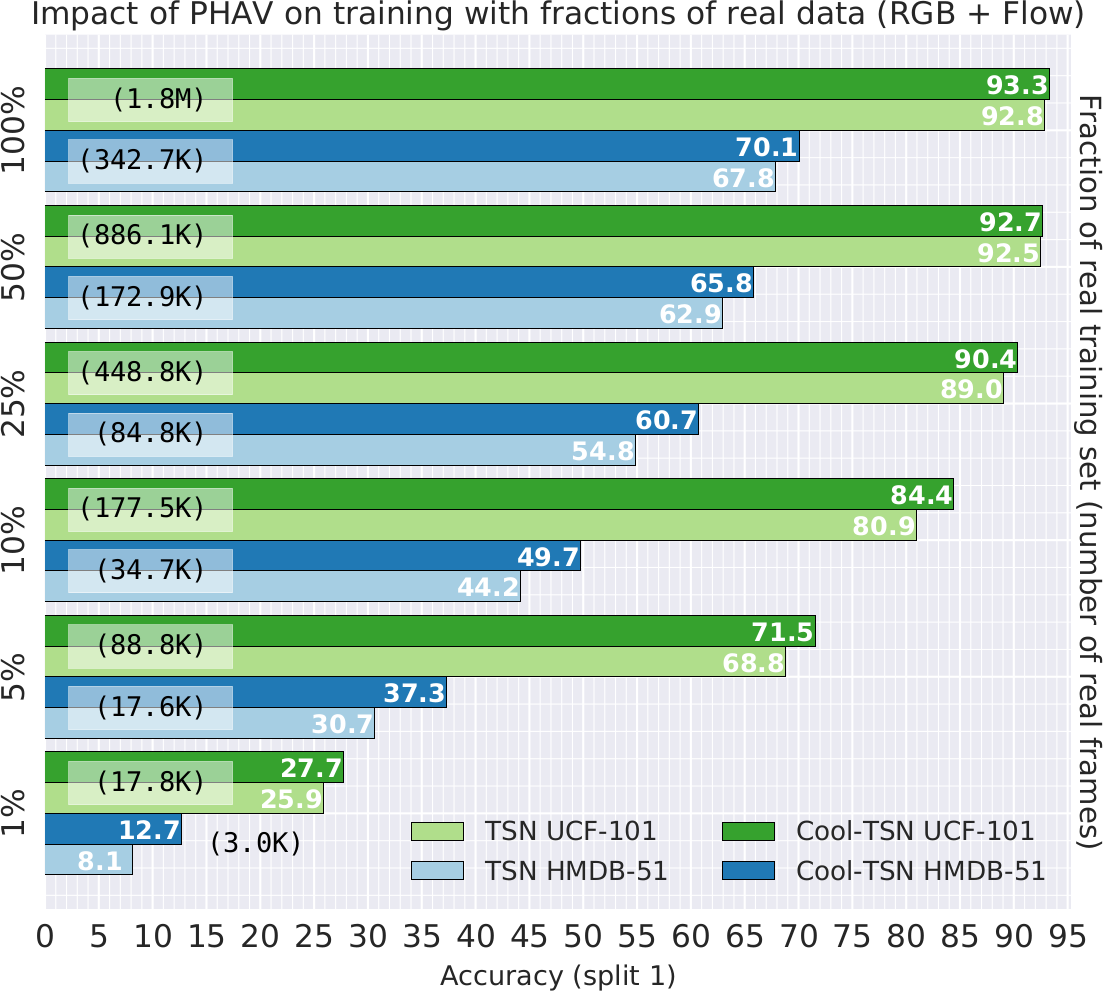}
        \caption{Impact of using subsets of the real world training videos
(split 1), with \vhad~(Cool-TSN) or without (TSN).}
        \label{fig:fractioning}
    \end{center}
\end{figure}

We further validate our hypothesis by investigating the impact of reducing the
number of real world training videos (and iterations), with or without the use
of \vhad.
Our results reported in Figure~\ref{fig:fractioning} confirms that
reducing training data from the target dataset impacts more severely TSN than Cool-TSN.
HMDB displays the largest gaps.
We partially attribute this to the smaller size of HMDB and also because some
categories of \vhad~overlap with some categories of HMDB.
Our results show that it is possible to replace half of HMDB with procedural
videos and still obtain comparable performance to using the full dataset
(65.8 vs. 67.8).
In a similar way, and although actions differ more, we show that reducing
UCF-101 to a quarter of its original training set still yields a Cool-TSN model
that rivals the state-of-the-art \cite{Wang2015,Simonyan2014,Wang2015d}.
This shows that our procedural generative model of videos can indeed be used to
augment different small real-world training sets and obtain better recognition
accuracy at a lower cost in terms of manual labor.

\subsection{Comparison with the state of the art}

In this section, we compare our model with the state of the art in action
recognition (Table \ref{table:sota}).
We separate the current state of the art into works that use one or multiple
sources of training data (such as by pre-training, multi-task learning or 
model transfer). We note that all works that use multiple sources can
potentially benefit from \vhad~without any modifications.
Our results indicate that our methods are competitive with the state of the art,
including methods that use much more manually labeled training data like the
Sports-1M dataset~\cite{Karpathy2014}.
Our approach also leads to better performance than the current best
generative video model VGAN~\cite{VondrickNIPS2016Generating} on UCF101, for
the same amount of manually labeled target real-world videos.
We note that while VGAN's more general task is quite challenging and different from ours, \cite{VondrickNIPS2016Generating} has also explored VGAN as a way to learn unsupervised representations useful for action recognition, thus enabling our comparison.

\begin{table}[]
    \vspace{-4mm}
    \small
    \centering
        \begin{tabularx}{\columnwidth}{p{4.0ex}lccc}
            \toprule
            &                                         & UCF-101             & HMDB-51             \\
            &  Method                                 & \%mAcc              & \%mAcc              \\
            \midrule                                 
\tabr{5}{\small\textsc{One source}}                                                                                                                                                   
            & iDT+FV \cite{Wang2013}                    & 84.8 &  57.2  \\ 
            & iDT+StackFV \cite{Peng2014b}              &  -                  &  66.8               \\ %
            & iDT+SFV+STP \cite{Wanga}                  & 86.0                &  60.1               \\ %
            & iDT+MIFS \cite{Lan2014}                   & 89.1                &  65.1               \\ %
            & VideoDarwin \cite{Fernando2015}    &  -                  &  63.7               &                 \\ 
            \midrule                                                                                                                                                                         
\tabr{9}{\textsc{Multiple sources}}                                                                                                                                                     
            & 2S-CNN \cite{Simonyan2014}                & 88.0                &  59.4               \\ %
            & TDD \cite{Wang2015d}                      & 90.3                &  63.2               \\ 
            & TDD+iDT \cite{Wang2015d}                  & 91.5                &  65.9               \\ 
            & C3D+iDT\cite{Tran2014}               & 90.4                &  -                  &                 \\ 
            & Actions$\sim$Trans \cite{Wang2015}        & 92.0                &  62.0               \\ 
            & 2S-Fusion \cite{Feichtenhofer2016}        & 93.5                &  69.2               \\ 
            & Hybrid-iDT \cite{DeSouza2016}             & 92.5                & 70.4                \\ 
            & TSN-3M \cite{Wangb}                       & 94.2                & 69.4                \\ 
            & VGAN \cite{VondrickNIPS2016Generating}    & 52.1                & -               \\ 
            \midrule
            & Cool-TSN                                   & 94.2                & 69.5                  \\
            \bottomrule
        \end{tabularx}
        \caption{Comparison against the state of the art.}
        \label{table:sota}
\end{table}



\section{Conclusion}\label{sec:conclusion}

In this work, we have introduced \vhad, a large synthetic dataset for action
recognition based on a procedural generative model of videos.
Although our model does not learn video representations like VGAN, it can
generate many diverse training videos thanks to its
grounding in strong prior physical knowledge about scenes, objects, lighting,
motions, and humans.

We provide quantitative evidence that our procedurally generated videos can be
used as a simple drop-in complement to small training sets of manually labeled
real-world videos.
Importantly, we show that we do not need to generate training videos for
particular target categories fixed a priori. Instead, surrogate categories
defined procedurally enable efficient multi-task representation learning for
potentially unrelated target actions that might have only few real-world
training examples.

Our approach combines standard techniques from Computer Graphics (procedural
generation) with state-of-the-art deep learning for action recognition. This
opens interesting new perspectives for video modeling and understanding,
including action recognition models that can leverage algorithmic ground truth
generation for optical flow, depth, semantic segmentation, or pose, or the
combination with unsupervised generative models like
VGAN~\cite{VondrickNIPS2016Generating} for dynamic background generation,
domain adaptation, or real-to-virtual world style transfer~\cite{Gatys2016}.

\ifcvprfinal
    \noindent \textbf{Acknowledgements.}
    Antonio M. Lopez is supported by the Spanish MICINN project TRA2014-57088-C2-1-R, 
    by the Secretaria d'Universitats i Recerca del Departament d'Economia i 
    Coneixement de la Generalitat de Catalunya (2014-SGR-1506), and the CERCA Programme~/~Generalitat de Catalunya. 
\fi

{\small
    \bibliographystyle{ieee}
    \bibliography{new}
}

\clearpage


\title{Procedural Generation of Videos to Train Deep Action Recognition Networks \\
    \vspace*{2mm}Supplementary material\vspace*{-4mm}}

\author{C\'{e}sar Roberto de Souza\textsuperscript{1,2}, Adrien Gaidon\textsuperscript{3}, Yohann Cabon\textsuperscript{1}, Antonio Manuel L\'{o}pez\textsuperscript{2} \\
    \textsuperscript{1}Computer Vision Group, NAVER LABS Europe, Meylan, France \\
    \textsuperscript{2}Centre de Visi\'{o} per Computador, Universitat Aut\`{o}noma de Barcelona, Bellaterra, Spain \\
    \textsuperscript{3}Toyota Research Institute, Los Altos, CA, USA \\
}

\maketitle
\setcounter{section}{0}

\vspace*{-5mm}


\section{Introduction}\label{sec:supp-intro}

This material provides additional information regarding our publication. In particular, we provide in-depth details about the parametric generative model we used to generate our procedural videos, an extended version of the probabilistic graphical model (whereas the graph shown in the publication had to be simplified due to size considerations), expanded generation statistics, details about additional data modalities we include in our dataset, and results for our Cool-TSN model for the separate RGB and flow streams.


\section{Generation details}\label{sec:supp-gen-model}

In this section, we provide more details about the
interpretable parametric generative model used in
our procedural generation of videos, presenting an
extended version of the probabilistic graphical model
given in our section 3.5.

\subsection{Variables}\label{ss:supp-variables}

We start by defining the main random variables used in 
our generative model. Here we focus only on critical variables
that are fundamental in understanding the orchestration of
the different parts of our generation, whereas all part-specific
variables are shown in Section \ref{ss:model}. The categorical
variables that drive most of the procedural generation are: 
\begin{equation}
\begin{aligned}
 H & : h &\in &~\{ {model}_1, {model}_2, \dots , {model}_{20} \}     \\
 A & : a &\in &~\{ ``clap", \dots, ``bump~into~each~other" \}             \\
 B & : b &\in &~\{ {motion}_1, {motion}_2, \dots , {motion}_{953} \} \\
 V & : v &\in &~\{ ``none", ``random~perturbation",                  \\
 &     &    &    ~~~``weakening", ``objects", ``blending" \}         \\
 C & : c &\in &~\{ ``kite", ``indoors", ``closeup", ``static" \}     \\
 E & : e &\in &~\{ ``urban", ``stadium", ``middle",                  \\
 &     &    &    ~~~``green", ``house", ``lake" \}                   \\ 
 D & : d &\in &~\{ ``dawn", ``day", ``dusk" \}                       \\ 
 W & : w &\in &~\{ ``clear", ``overcast", ``rain, ``fog" \}          \\
 \end{aligned}
\end{equation}

\noindent where
$H$ is the human model to be used by the protagonist,
$A$ is the action category to be generated,
$B$ is the base motion sequence used for the action,
$V$ is the variation to be applied to the base motion, 
$C$ is the camera behavior,
$E$ is the environment of the virtual world where the action will take place,
$D$ is the day phase, 
$W$ is the weather condition.

%

These categorical variables are in turn controlled by a group
of parameters that can be adjusted in order to drive the sample
generation. These parameters include the $\theta_A$ parameters of
a categorical distribution on action categories $A$, 
the $\theta_W$ for weather conditions $W$,
 $\theta_D$ for day phases $D$,
 $\theta_H$ for model models $H$,
 $\theta_V$ for variation types $V$, and
 $\theta_C$ for camera behaviors $C$.

Additional parameters include the conditional probability tables 
of the dependent variables:
a matrix of parameters $\theta_{AE}$ where each row contains
    the parameters for categorical distributions on
    environments $E$ for each action category $A$,
the matrix of parameters $\theta_{AC}$ on camera behaviors $C$ for each action $A$,
the matrix of parameters $\theta_{EC}$ on camera behaviors $C$ for each environment $E$, and
the matrix of parameters $\theta_{AB}$ on motions $B$ for each action $A$.

Finally, other relevant parameters include
$T_{min}$, $T_{max}$, and $T_{mod}$, the minimum, maximum and most likely 
durations for the generated video. We denote the set of all parameters
in our model by $\bm{\theta}$.

\subsection{Model} \label{ss:model}

\begin{figure}
    \center
    \includegraphics[width=1.0\columnwidth]{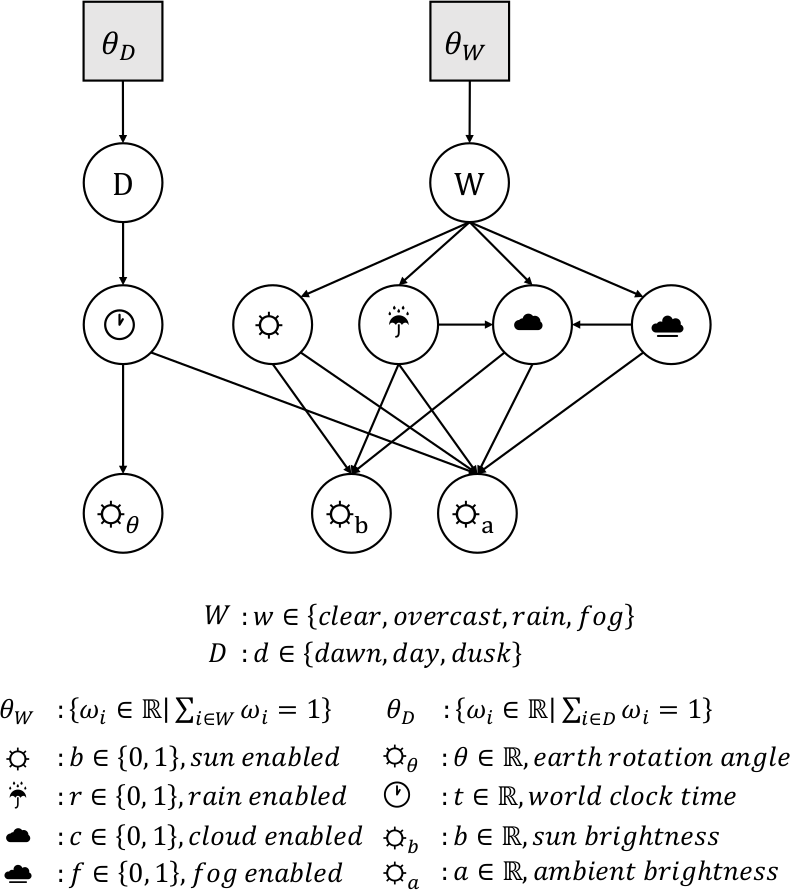}
    \caption{Probabilistic graphical model for $P_1(D, W \mid \bm{\theta})$,
        the first part of our parametric generator (world time and weather).}
    \label{fig:pgm1}
    \vspace{-3mm}
\end{figure}

\begin{figure}
    \center
    \includegraphics[width=0.65\columnwidth]{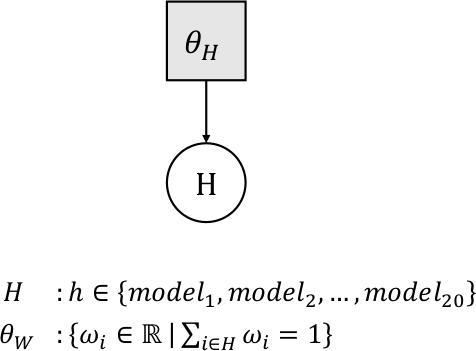}
    \caption{Probabilistic graphical model for $P_2(H \mid \bm{\theta})$,
        the second part
        of our parametric generator (human models).}
    \label{fig:pgm2}
\end{figure}

\begin{figure*}
    \center
    \includegraphics[width=\textwidth]{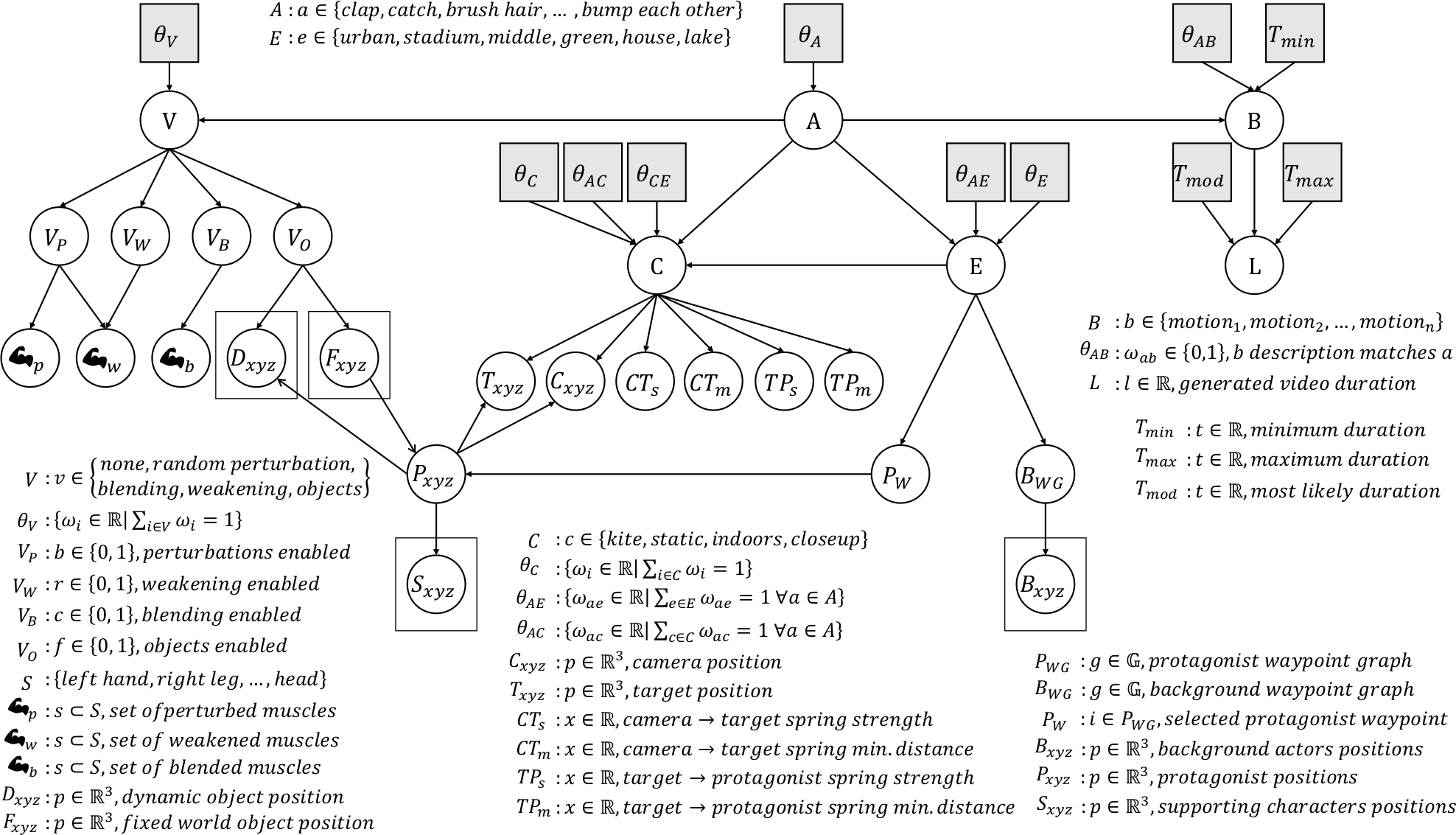}
    \caption{Probabilistic graphical model for $P_3(A, L, B, V, C, E, W \mid \bm{\theta})$,
        the third part of our parametric generator (scene and 
        action preparation).}
    \label{fig:pgm3}
\end{figure*}

The complete interpretable parametric probabilistic model used by
our generation process, given our generation parameters $\bm{\theta}$,
can be written as:
\begin{equation}
\begin{aligned}
P(H,&A, L, B, V, C, E, D, W \mid \bm{\theta}) = \\
    &~~ P_1(D, W \mid \bm{\theta})~P_2(H \mid \bm{\theta}) \\
    &~~ P_3(A, L, B, V, C, E, W \mid \bm{\theta}) \\
\end{aligned}
\end{equation}
\noindent where $P_1$, $P_2$ and $P_3$ are defined by the probabilistic
graphical models represented on Figure \ref{fig:pgm1}, \ref{fig:pgm2}
and \ref{fig:pgm3}, respectively. We use extended plate 
notation~\cite{Bishop2006} to indicate repeating variables, marking 
parameters (non-variables) using filled rectangles.

\subsection{Distributions}\label{ss:distributions}

The generation process makes use of four main families of
distributions: categorical, uniform, Bernoulli and triangular.
We adopt the following three-parameter formulation for the
triangular distribution:
\begin{equation}
Tr(x \mid a, b, c) = \begin{cases}
  0                         & \text{for } x < a, \\
  \frac{2(x-a)}{(b-a)(c-a)} & \text{for } a \le x < c, \\[4pt]
  \frac{2}{b-a}             & \text{for } x = c,       \\[4pt]
  \frac{2(b-x)}{(b-a)(b-c)} & \text{for } c < x \le b, \\[4pt]
  0                         & \text{for } b < x.
  \end{cases}
\end{equation}

\noindent All distributions are implemented using the open-source 
Accord.NET Framework\footnote{\url{http://accord-framework.net}}. While 
we have used mostly uniform distributions to create the dataset used
in our experiments, we have the possibility to bias the generation
towards values that are closer to real-world dataset statistics.

\noindent \textbf{Day phase.}
As real-world action recognition datasets are more likely to contain video recordings captured during daylight, we fixed the parameter $\theta_D$ such that
\begin{equation}
\begin{aligned}
    P(D =&~dawn  &\mid \theta_D)   &=& \nicefrac{1}{3}& \\
    P(D =&~day   &\mid \theta_D)   &=& \nicefrac{1}{3}& \\
    P(D =&~dusk  &\mid \theta_D)   &=& \nicefrac{1}{3}& \\
    P(D =&~night &\mid \theta_D)   &=& 0&.   \\
\end{aligned}
\end{equation}

\noindent We note that although our system can also generate
night samples, we do not include them in \vhad~at this moment.

\noindent \textbf{Weather.}
In order to support a wide range of applications of our 
dataset, we fixed the parameter $\theta_W$ such that 

\begin{equation}
\begin{aligned}
P(W =&~clear    &\mid \theta_W)   &=& \nicefrac{1}{4}& \\
P(W =&~overcast &\mid \theta_W)   &=& \nicefrac{1}{4}& \\
P(W =&~rain     &\mid \theta_W)   &=& \nicefrac{1}{4}& \\
P(W =&~fog      &\mid \theta_W)   &=& \nicefrac{1}{4}&, \\
\end{aligned}
\end{equation}

\noindent ensuring all weather conditions are present.

\noindent \textbf{Camera.}
In addition to the Kite camera, we also included 
specialized cameras that can be enabled only for certain
environments (Indoors), and certain actions (Close-Up). 
We fixed the parameter $\theta_C$ such that

\begin{equation}
\begin{aligned}
P(C =&~kite    &\mid \theta_C)   &=& \nicefrac{1}{3}& \\
P(C =&~closeup &\mid \theta_C)   &=& \nicefrac{1}{3}& \\
P(C =&~indoors &\mid \theta_C)   &=& \nicefrac{1}{3}&. \\
\end{aligned}
\end{equation}

\noindent However, we have also fixed $\theta_{CE}$ and $\theta_{AC}$ 
such that the Indoors camera is only available for the
house environment, and that the Close-Up camera can also
be used for the \textit{BrushHair} action in addition to Kite.

\noindent \textbf{Environment, human model and variations.} 
We fixed the parameters $\theta_E$, $\theta_H$, and $\theta_V$  
using equal weights, such that the variables $E$, $H$, and $V$
can have uniform distributions.

\noindent \textbf{Base motions.} 
All base motions are weighted according to the 
minimum video length parameter $T_{min}$, where
motions whose duration is less than $T_{min}$
are assigned weight zero, and others are set
to uniform, such that
\begin{equation}
P(B = b | T_{min}) \propto \begin{cases}
                                1 & \text{if } length(b) \ge T_{min} \\
                                0 & \text{otherwise} \\
                            \end{cases} 
\end{equation}

\noindent We then perform the selection of a motion $B$ given a category
$A$ by introducing a list of regular expressions associated with each
of the action categories. We then compute matches between the textual
description of the motion in its source (\eg short text descriptions
in \cite{CarnegieMellonGraphicsLab2016}) and these expressions, such
that
\begin{equation}
\begin{aligned}
 (\theta_{AB})_{ab}= \begin{cases}
                        1 & \text{if } match(\text{regex}_{a}, \text{desc}_b) \\
                        0 & \text{otherwise} \\
\end{cases} \forall a \in A, \forall b \in B. \\
\end{aligned}
\end{equation}

\noindent We then use $\theta_{AB}$ such that

\begin{equation}
P(B = b \mid A = a, \theta_{AB}) \propto (\theta_{AB})_{a,b}. \\
\end{equation}

\noindent \textbf{Weather elements.}
The selected weather $W$ affects world parameters such as the
sun brightness, ambient luminosity, and multiple boolean 
variables that control different aspects of the world
(\cf Figure \ref{fig:pgm1}). The activation of one of these 
boolean variables (\eg fog visibility) can influence the
activation of others (\eg clouds) according to Bernoulli
distributions ($p=0.5$).

\noindent \textbf{World clock time.}
The world time is controlled depending on $D$. In order to avoid
generating a large number of samples in the borders between two
periods of the day, where the distinction between both phases is
blurry, we use different triangular distributions associated with
each phase, giving a larger probability to hours of interest (sunset,
dawn, noon) and smaller probabilities to hours at the
transitions. We therefore define the distribution of the world
clock times $P(T)$ as:

\begin{equation}
P(T = t \mid D) \propto \sum_{d \in D} P(T = t \mid D = d)
\end{equation}

\noindent where
\begin{equation}
\begin{aligned}
P(T = t \mid D = &dawn&) = Tr( t \mid&~7h,&10h,~&9h&)&  \\
P(T = t \mid D = &day&)  = Tr( t \mid&~10h,&16h,~&13h&)&  \\
P(T = t \mid D = &dusk&) = Tr( t \mid&~17h,&20h,~&18h&)&. \\
\end{aligned}
\end{equation}

\noindent \textbf{Generated video duration.}
The selection of the clip duration $L$ given the selected
motion $b$ is performed considering the motion length $L_b$,
the maximum video length $T_{min}$ and the desired mode $T_{mod}$:
\begin{equation}
\begin{aligned}
  P(L = l \mid B = b) = Tr(&a=T_{min}, \\
                           &b=min(L_b, T_{max}), \\
                           &c = min(T_{mod}, L_b)) \\
\end{aligned}
\end{equation}

\begin{figure*}[]
 \begin{center}
     \begin{subfigure}{0.47\textwidth}
         \includegraphics[width=\textwidth,trim={0 2cm 0 2.31cm},clip]{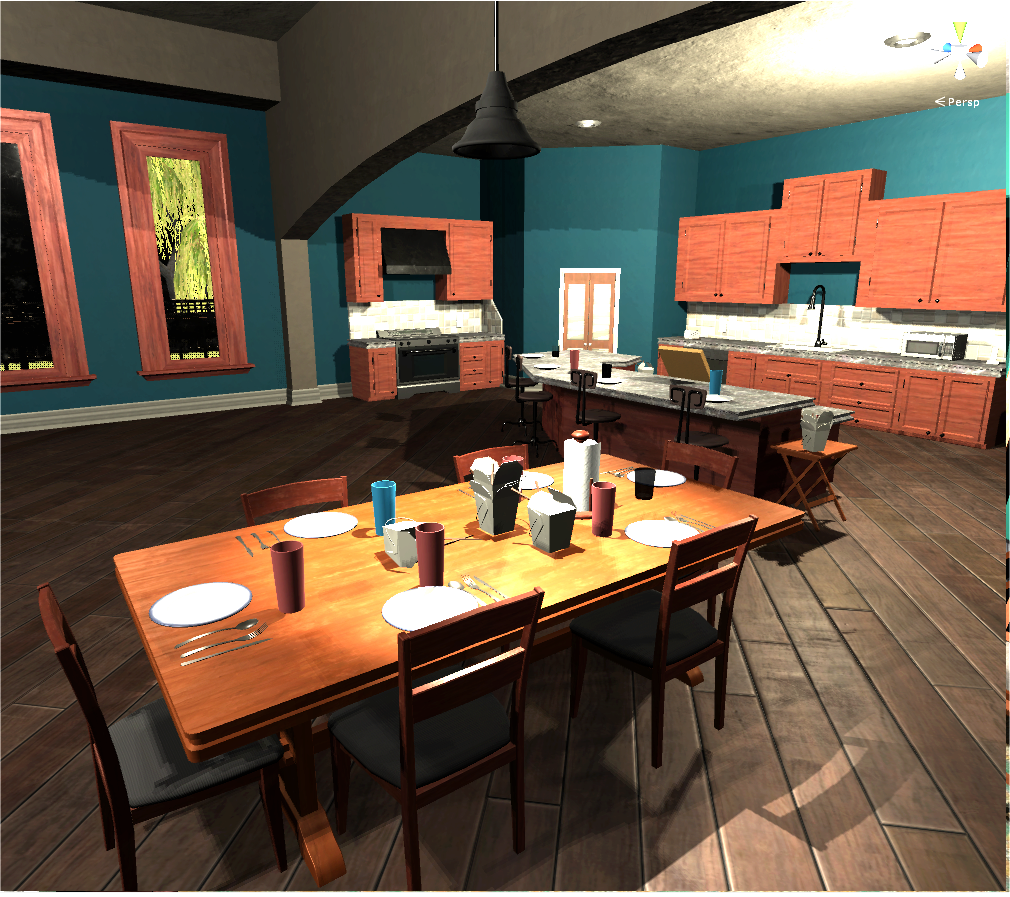}
        \end{subfigure}
        \begin{subfigure}{0.47\textwidth}
            \includegraphics[width=\textwidth,trim={0 0 0 0},clip]{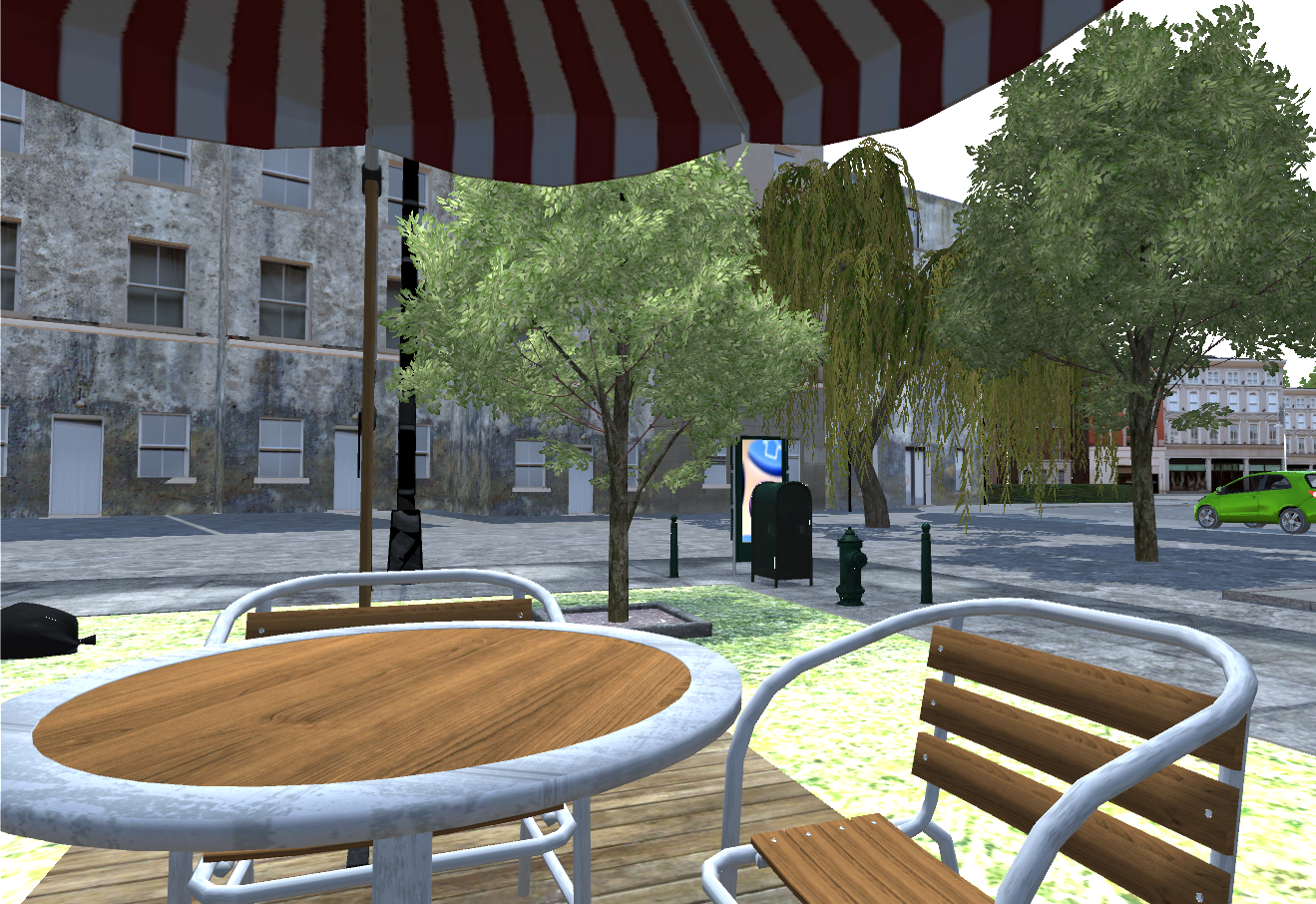}
        \end{subfigure}
    \end{center}
    \caption{Example of indoor and outdoors scenes.}
    \label{fig:short}
\end{figure*}

\noindent \textbf{Actors placement and environment.} Each
environment $E$ has at most two associated waypoint graphs.
One graph refers to possible positions for the protagonist, 
while an additional second graph gives possible positions $B_{WG}$ for
spawning background actors. Indoor scenes (\cf Figure \ref{fig:short}) 
do not include background actor graphs.
After an environment has been selected, a waypoint $P_W$ is randomly
selected from the graph using an uniform distribution. The protagonist
position $P_{xyz}$ is then set according to the position of $P_W$.
The $S_{xyz}$ position of each supporting character, if any, is set 
depending on $P_{xyz}$. The position and destinations for the
background actors are set depending on $B_{WG}$.

\noindent \textbf{Camera placement and parameters.} After a camera
has been selected, its position $C_{xyz}$ and the position $T_{xyz}$ 
of the target are set depending on the position $P_{xyz}$ of the 
protagonist. The camera parameters are randomly sampled using uniform
distributions on sensible ranges according to the observed behavior
in Unity. The most relevant secondary variables for the camera are
shown in Figure \ref{fig:pgm3}. They include Unity-specific parameters
for the camera-target ($CT_s$, $CT_m$) and target-protagonist springs
($TP_s$, $CT_m$) that can be used to control their strength and a
minimum distance tolerance zone in which the spring has no effect 
(remains at rest). In our generator, the minimum distance is set
to either 0, 1 or 2 meters with uniform probabilities. This setting
is responsible for a "delay" effect that allows the protagonist to 
not be always in the center of camera focus (and thus avoiding
creating such bias in the data).

\begin{figure}[]
    \begin{center}
        \vspace*{1mm}
        \includegraphics[width=0.86\columnwidth]{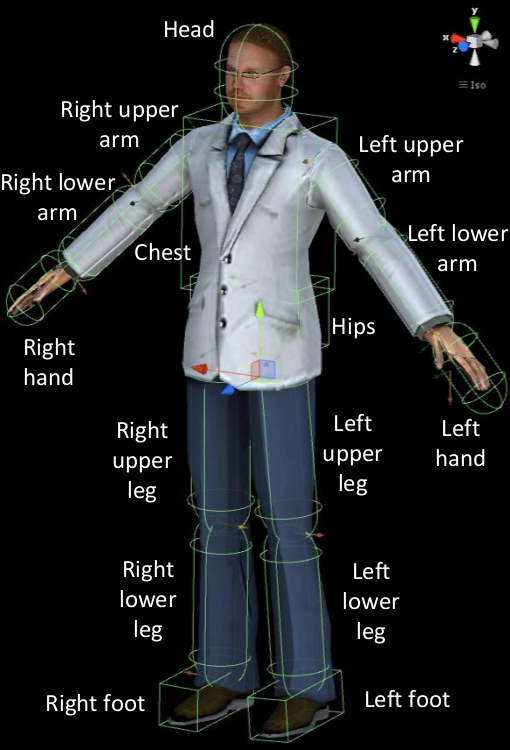}
        \caption{Ragdoll configuration with 15 muscles.}
        \label{fig:ragdoll}
        \vspace*{-8mm}
    \end{center}
\end{figure}

\noindent \textbf{Action variations.} After a variation mode has
been selected, the generator needs to select a subset of the
ragdoll muscles (\cf Figure \ref{fig:ragdoll}) to be perturbed
(random perturbations) or to be replaced with movement from a 
different motion (action blending). 
These muscles are selected using a uniform distribution on muscles
that have been marked as non-critical depending on the previously
selected action category $A$. When using weakening, a subset of 
muscles will be chosen to be weakened with varying parameters 
independent of the action category. When using objects, the choice
of objects to be used and how they have to be used is also dependent
on the action category.

\noindent \textbf{Object placement.} Interaction with objects can happen
in two forms: dynamic or static. When using objects dynamically,
an object of the needed type (\eg bow, ball) is spawned around (or is attached
to) the protagonist at a pre-determined position, and is manipulated using
3D joints, inverse kinematics, or both. When using static (fixed) objects,
the protagonist is moved to the vicinity of an object already present in
the virtual world (\eg bench, stairs).

\subsection{Statistics}\label{ss:statistics}

\begin{figure}[]
    \center
    \includegraphics[width=\columnwidth]{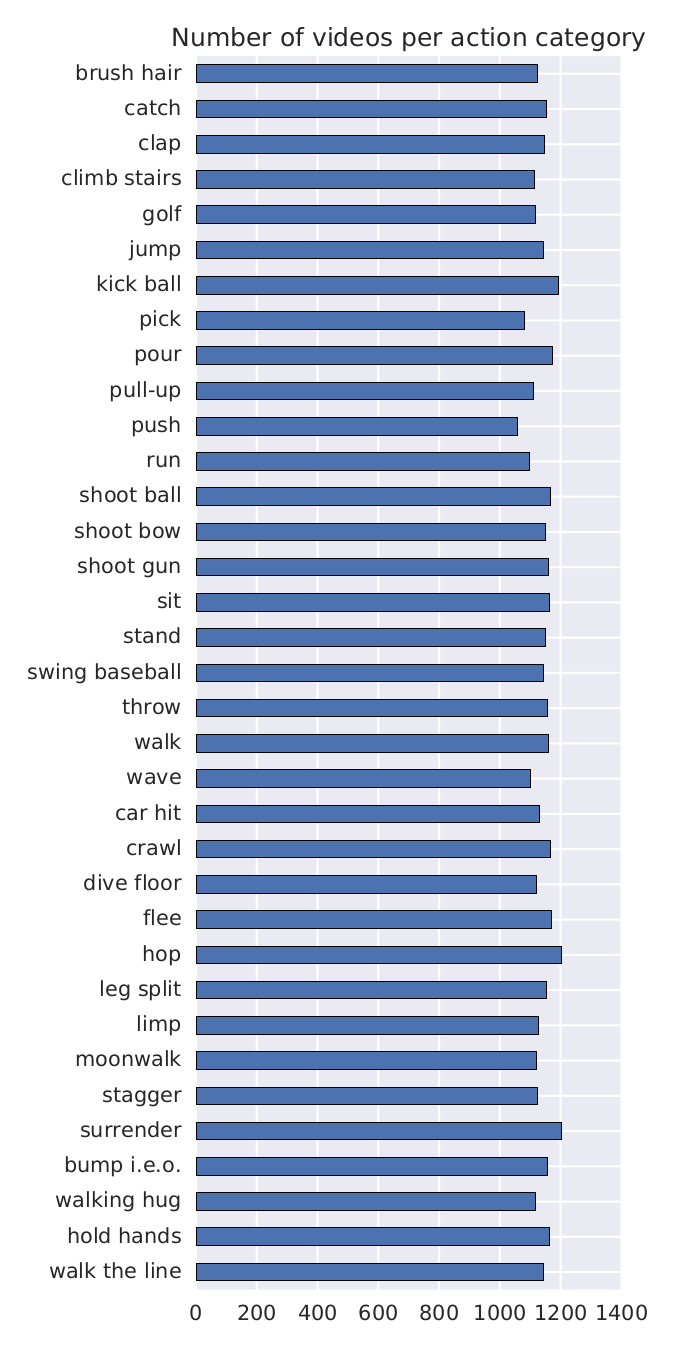}
    \caption{Plot of the number of videos generated for each category in
        the version of our PHAV dataset used in the publication.}
    \label{fig:stats-full}
\end{figure}

In this section we show statistics about the version of PHAV
that has been used in experimental section of our paper.
A summary of the key statistics for the generated dataset can be seen in Table \ref{statistics}. 
Figure \ref{fig:stats-full} shows the number of videos generated
after each action category in PHAV. As it can be seen, the number
is higher than 1,000 samples for all categories.

\begin{figure}[]
    \center
    \includegraphics[width=\columnwidth,trim={0 0 0 0},clip]{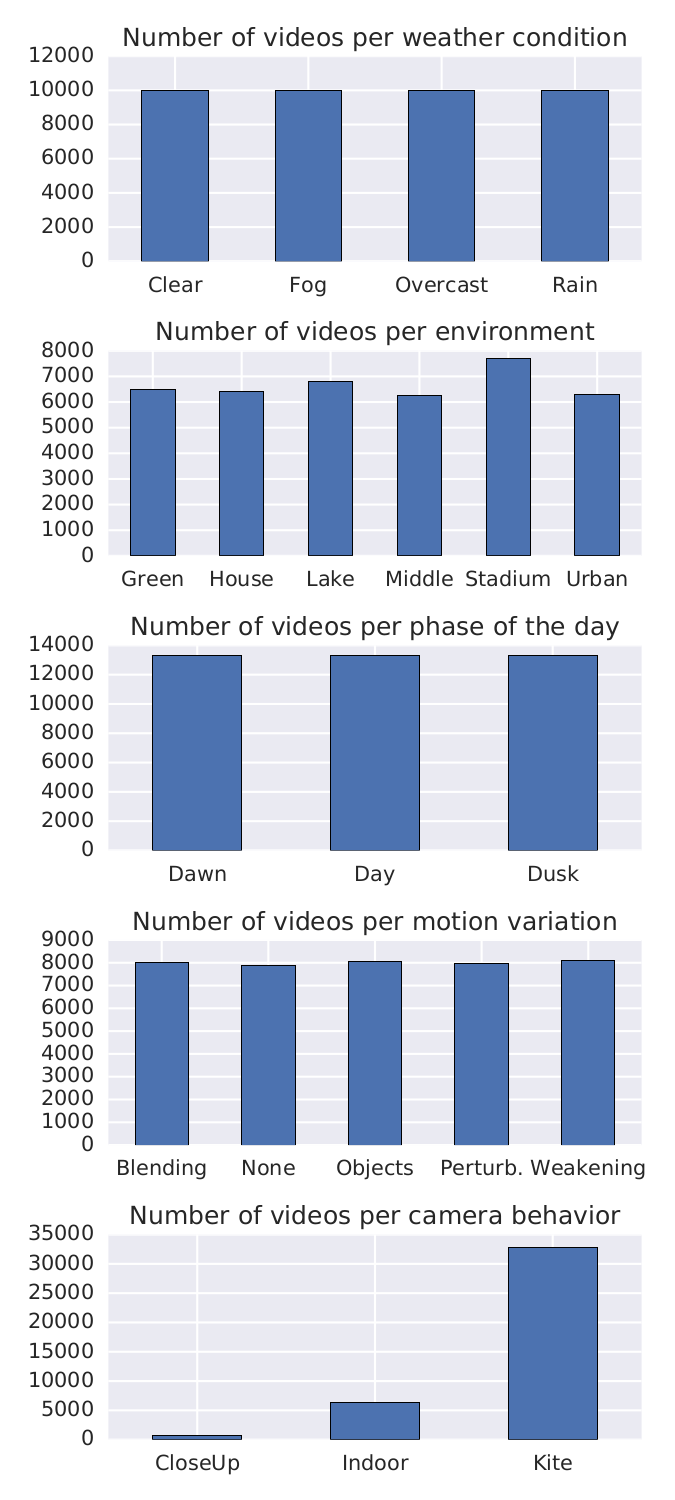}
    \vspace{-5mm}
    \caption{Plot of the number of videos per parameter value.}
    \label{fig:stats-parameter}
    \vspace{-7mm}
\end{figure}

Figure \ref{fig:stats-parameter} shows the number of videos generated
by value of each main random generation variable. The histograms 
reflect the probability values presented in Section~\ref{ss:distributions}.
While our parametric model is flexible enough to generate a wide range 
of world variations, we have focused on generating videos that would 
be more similar to those in the target datasets.

\begin{table}[]
    \begin{center}
        \begin{tabular}{lc}
            \toprule
            Statistic & Value \\
            \midrule
            Clips                      & 39,982    \\
            Total dataset frames       & 5,996,286 \\
            Total dataset duration     & 2d07h31m  \\
            Average video duration     & 4.99s     \\
            Average number of frames   & 149.97    \\
            Frames per second          & 30        \\
            Video dimensions           & 340x256   \\
            Average clips per category & 1,142.3   \\
            Image modalities (streams) & 6         \\
            \bottomrule
        \end{tabular}
    \end{center}
    \vspace*{-5mm}
    \caption{Statistics of the generated dataset instance. }
    \label{statistics}
    \vspace*{-5mm}
\end{table}

\subsection{Data modalities}\label{ss:modalities}

Although not discussed in the paper, our generator can
also output multiple data modalities for a single video,
which we include in our public release of PHAV.
Those data modalities are rendered roughly at the same time using
Multiple Render Targets (MRT), resulting in a superlinear speedup 
as the number of simultaneous output data modalities grow.
The modalities in our public release include:

\begin{figure}[htpb]
    \begin{center}
        \begin{subfigure}{0.4951\linewidth}
            \includegraphics[width=\textwidth]{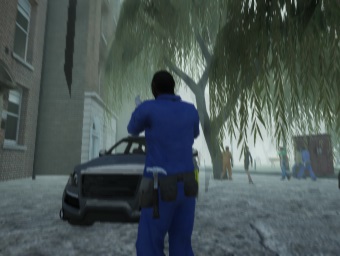}
        \end{subfigure}
        \begin{subfigure}{0.4951\linewidth}
            \includegraphics[width=\textwidth]{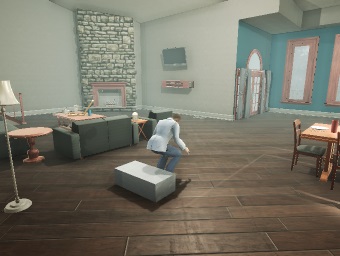}
        \end{subfigure}
        
        \begin{subfigure}{0.4951\linewidth}
            \includegraphics[width=\textwidth]{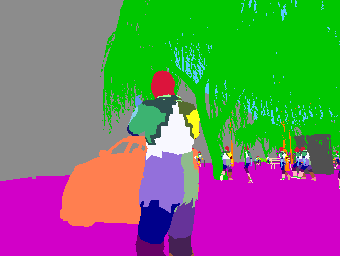}
        \end{subfigure}
        \begin{subfigure}{0.4951\linewidth}
            \includegraphics[width=\textwidth]{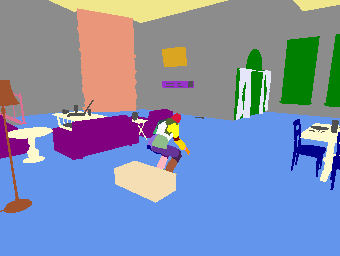}
        \end{subfigure}
        
        \begin{subfigure}{0.4951\linewidth}
            \includegraphics[width=\textwidth]{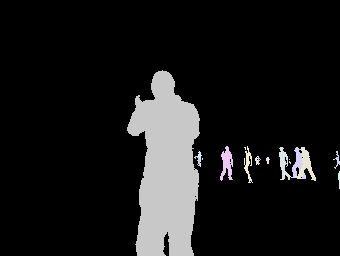}
        \end{subfigure}
        \begin{subfigure}{0.4951\linewidth}
            \includegraphics[width=\textwidth]{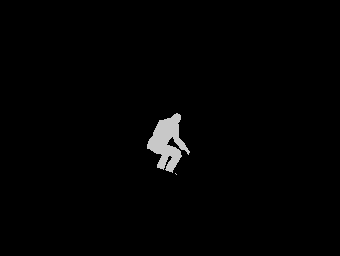}
        \end{subfigure}
        
        \begin{subfigure}{0.4951\linewidth}
            \includegraphics[width=\textwidth]{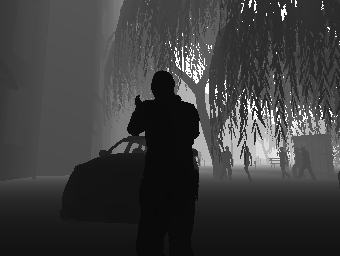}
        \end{subfigure}
        \begin{subfigure}{0.4951\linewidth}
            \includegraphics[width=\textwidth]{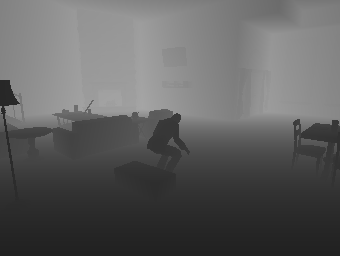}
        \end{subfigure}
        
        \begin{subfigure}{0.4951\linewidth}
            \includegraphics[width=\textwidth]{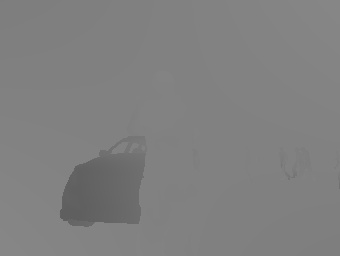}
        \end{subfigure}
        \begin{subfigure}{0.4951\linewidth}
            \includegraphics[width=\textwidth]{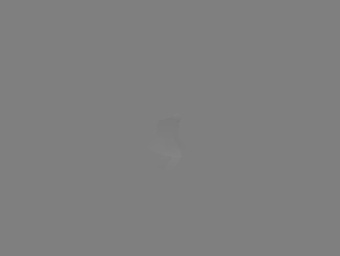}
        \end{subfigure}
        
        \begin{subfigure}{0.4951\linewidth}
            \includegraphics[width=\textwidth]{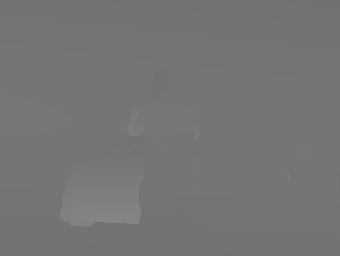}
        \end{subfigure}
        \begin{subfigure}{0.4951\linewidth}
            \includegraphics[width=\textwidth]{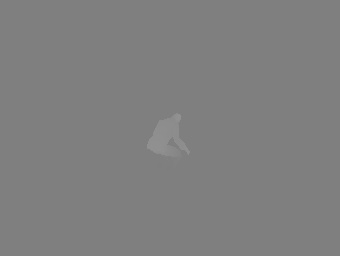}
        \end{subfigure}
    \end{center}    
    \caption{Example frames and data modalities for a synthetic action 
        (car hit, left) and MOCAP-based action (sit, right). From top
        to bottom: Rendered RGB Frames, Semantic Segmentation, Instance Segmentation, Depth Map, Horizontal Optical Flow, and Vertical Optical Flow. Depth image brightness has been adjusted in this figure to ensure visibility on paper.}
    \label{fig:modalities}
\end{figure}

\noindent \textbf{Rendered RGB Frames.} 
Those are the RGB frames that constitute the action video. They
are rendered at 340x256 resolution and 30 FPS such that they can
be directly feed to Two-Stream style networks. Those frames have been
post-processed with 2x Supersampling Anti-Aliasing (SSAA), motion blur, 
bloom, ambient occlusion, screen space reflection, color grading, and
vignette.

\noindent \textbf{Semantic Segmentation.} 
Those are the per-pixel semantic segmentation ground-truths containing the object class label annotations for every pixel in the RGB frame. They are encoded as sequences of 24-bpp PNG files with the same resolution as the RGB frames. We provide 63 pixel classes, including the same 14 classes used in Virtual KITTI \cite{Gaidon2016}, classes specific for indoor scenarios, classes for dynamic objects used in every action, and 27 classes depicting body joints and limbs.

\noindent \textbf{Instance Segmentation.} 
Those are the per-pixel instance segmentation ground-truths containing
the person identifier encoded as different colors in a sequence of frames.
They are encoded in exactly the same way as the semantic segmentation
ground-truth explained above.

\noindent \textbf{Depth Map.} 
Those are depth map ground-truths for each frame. They are represented as a sequence of 16-bit grayscale PNG images with a fixed far plane of 655.35 meters. This encoding ensures that a pixel intensity of 1 can correspond to a 1cm distance from the camera plane.

\noindent \textbf{Optical Flow.} 
Those are the ground-truth (forward) optical flow fields computed from 
the current frame to the next frame. We provide separate sequences of
frames for the horizontal and vertical directions of optical flow
represented as sequences of 16-bpp JPEG images with the same resolution
as the RGB frames.

\noindent \textbf{Raw RGB Frames.} 
Those are the raw RGB frames before any of the post-processing effects
mentioned above are applied. This modality is mostly included for completeness,
and has not been used in experiments shown in the paper.

\noindent \textbf{Pose, location and additional information.}
Although not an image modality, our generator can also produce textual annotations for every frame. Annotations include camera parameters, 3D and 2D bounding boxes, joint locations in screen coordinates (pose), and muscle information (including muscular strength, body limits and other physical-based annotations) for every person in a frame.


\section{Experiments}\label{sec:supp-experiments}

In this section, we show more details about the experiments
shown in the experimental section of our paper.

Table \ref{results_frac} shows the impact of training our
Cool-TSN models using only a fraction of the real world data
(Figure 7 of original publication) in a tabular format. As
it can be seen, mixing real-world and virtual-world data
from \vhad is helpful in almost all cases.

\begin{figure}
    \center
    \begin{subfigure}{\columnwidth}
        \includegraphics[width=\columnwidth,trim={0cm 5mm 0cm 0cm},clip]{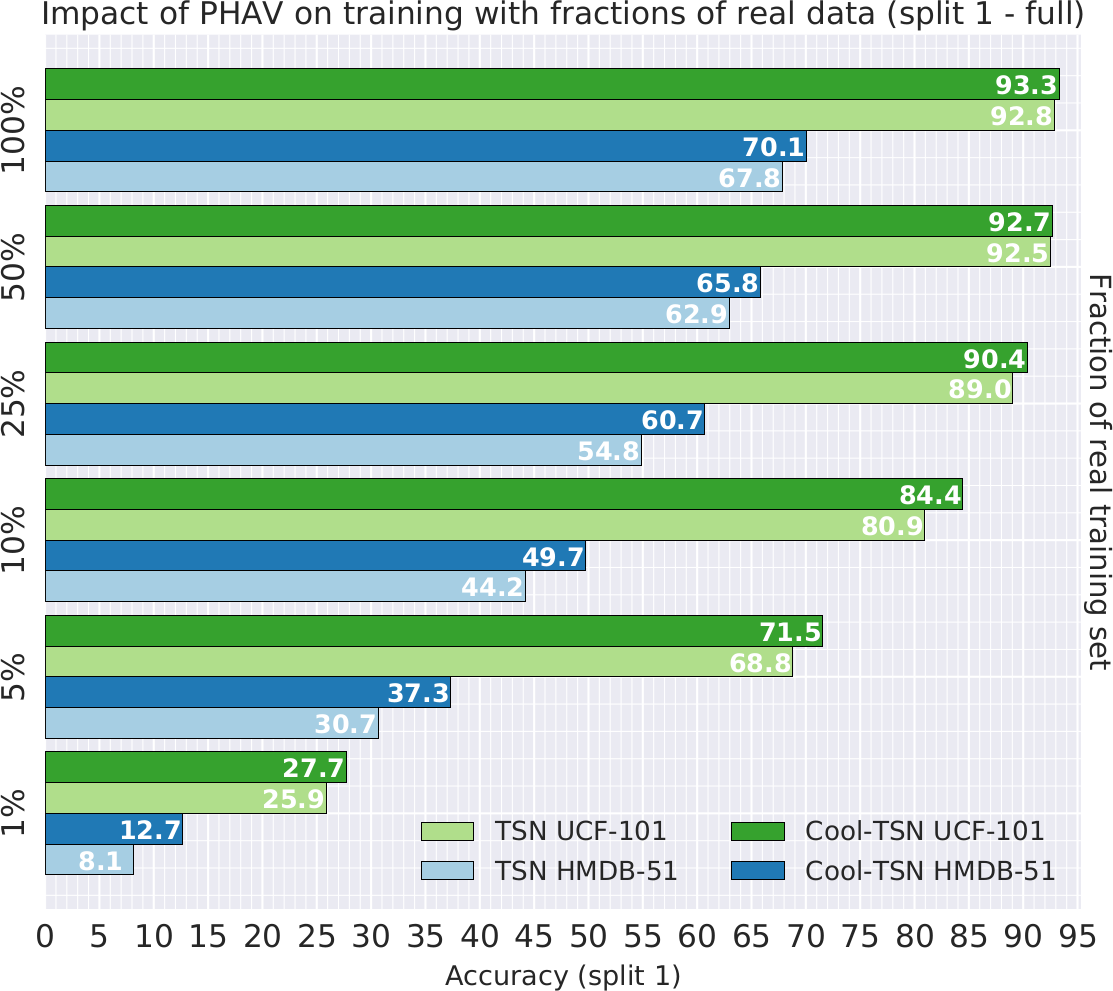}
    \end{subfigure}
    \begin{subfigure}{\columnwidth}
        \includegraphics[width=\columnwidth,trim={0cm 5mm 0cm -5mm },clip]{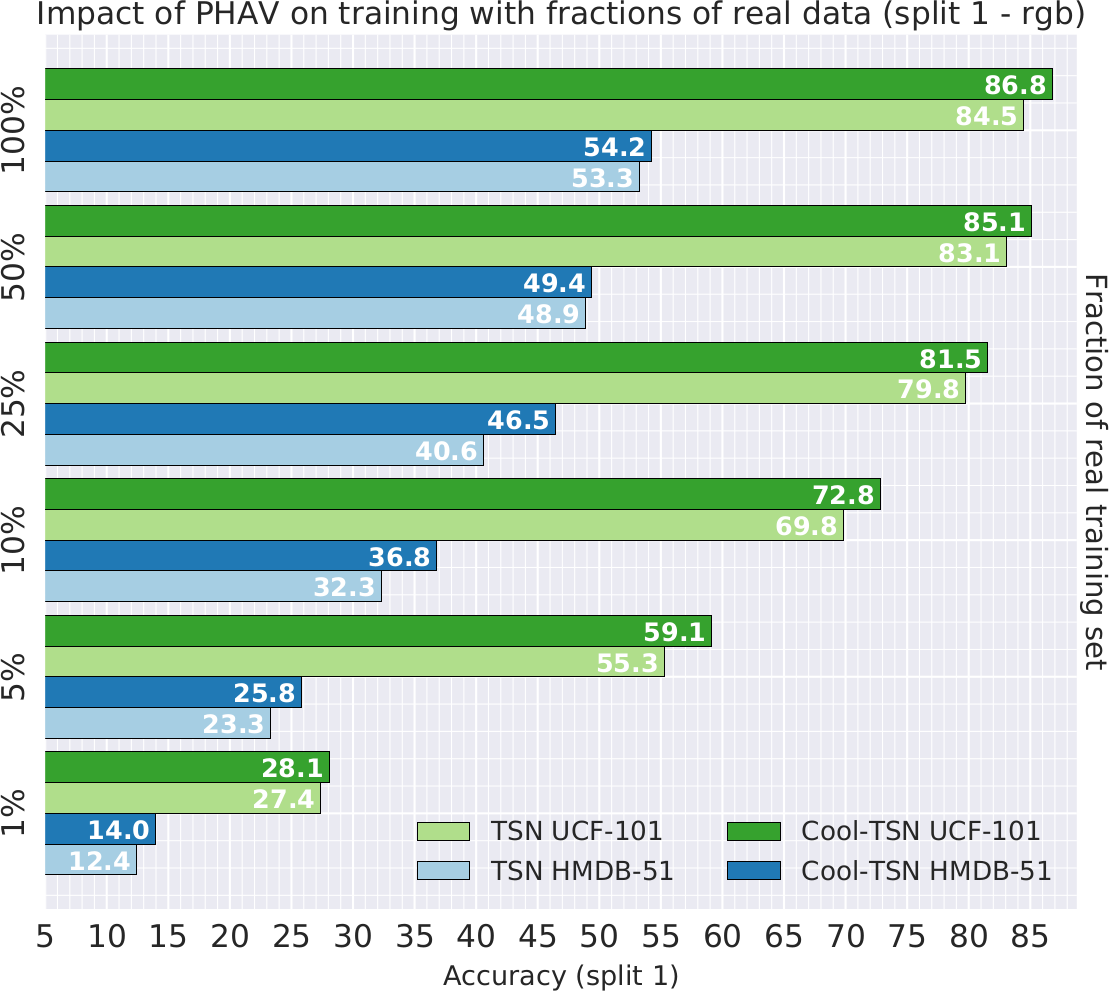}
    \end{subfigure}
    \begin{subfigure}{\columnwidth}
        \includegraphics[width=\columnwidth,trim={0cm 0cm 0cm -5mm },clip]{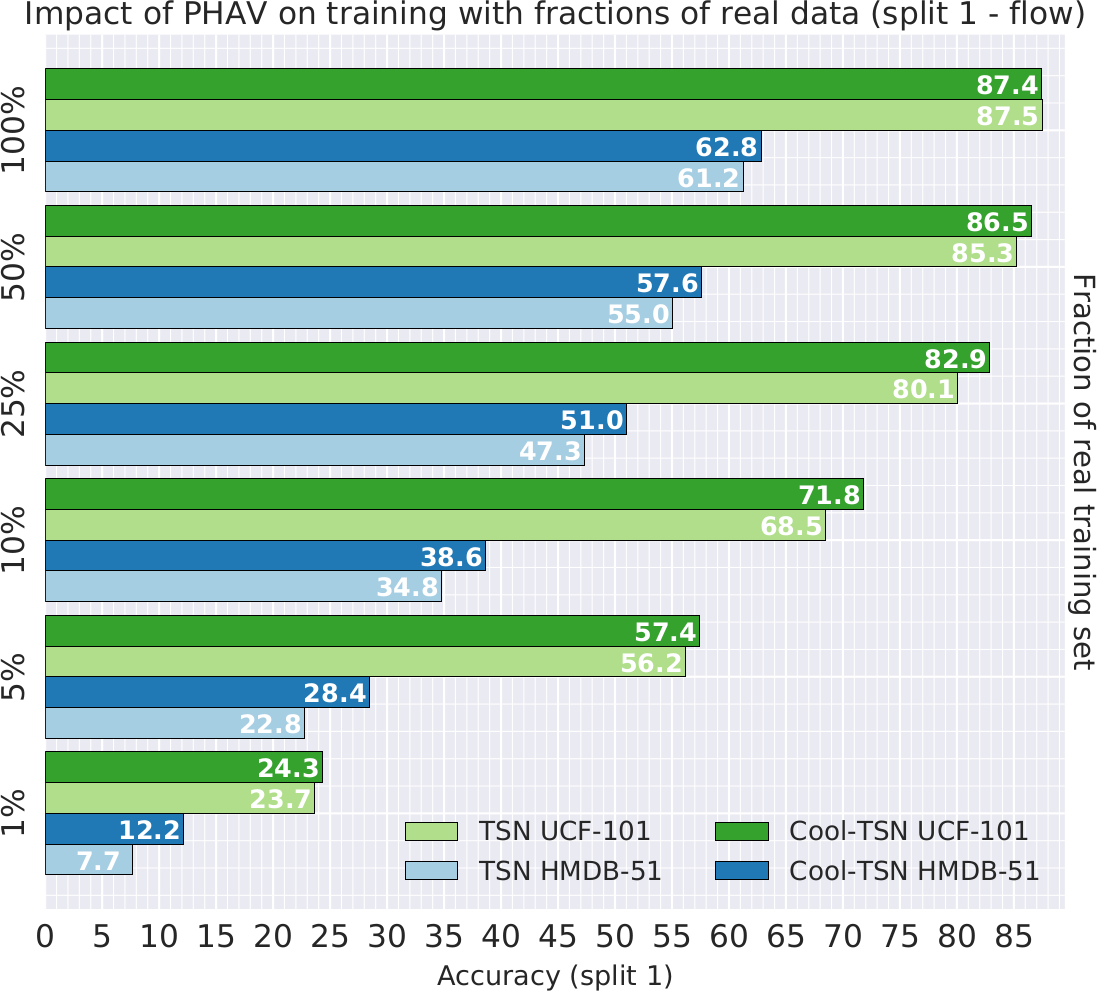}
    \end{subfigure}
    \vspace*{-2mm}
    \caption{TSN and Cool-TSN results for different amounts of
        training data for combination and separate streams.}
    \label{fig:fract}
\end{figure}

Figure \ref{fig:fract} shows the performance of each network
stream separately. The second image on the row shows the
performance on the Spatial (RGB) stream. The last image on
the row shows the performance for the Temporal (optical flow)
stream.
One can see how the optical flow stream is the biggest responsible
for the good performance of our Cool-TSN, including when using very
low fractions of the real data. This confirms that our generator 
is indeed producing plausible motions that are being helpful to
learn both the virtual and real-world data sources.

 \begin{table}[]
     \centering
     \begin{tabular}{@{}c|cc|cc@{}}
         \toprule
         \footnotesize{Fraction} & \footnotesize{UCF101} & \footnotesize{UCF101+\vhad} & \footnotesize{HMDB51} & \footnotesize{HMDB51+\vhad} \\
         \midrule 
         1\%  & 25.9  & \textbf{27.7}  &  8.1   & \textbf{12.7}   \\
         5\%  & 68.5  & \textbf{71.5}  & 30.7   & \textbf{37.3}   \\
         10\% & 80.9  & \textbf{84.4}  & 44.2   & \textbf{49.7}   \\
         25\% & 89.0  & \textbf{90.4}  & 54.8   & \textbf{60.7}   \\
         50\% & 92.5  & \textbf{92.7}  & 62.9   & \textbf{65.8}   \\
        100\% & 92.8  & \textbf{93.3}  & 67.8   & \textbf{70.1}   \\
     \bottomrule
     \end{tabular}
     \caption{TSN and Cool-TSN (+\vhad) with different
         fractions of real-world training data (split 1).}
     \label{results_frac}
 \end{table}

\section{Video}\label{sec:video}

\begin{figure*}[p]
    \center
    \includegraphics[width=0.8\textwidth]{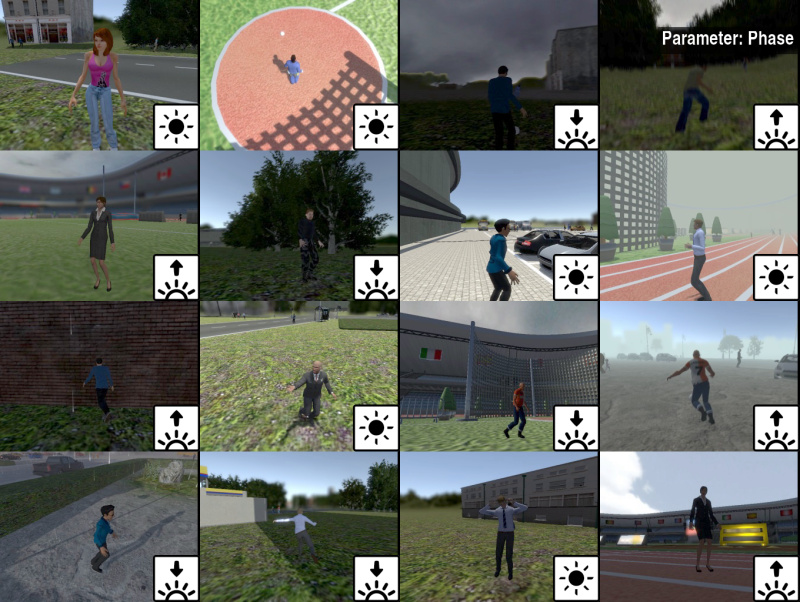}
    \caption{Sample frame from the supplementary video
        available at \url{http://adas.cvc.uab.es/phav/}.}
    \label{fig:video}
\end{figure*}

\begin{figure*}[p]
    \center
    \includegraphics[width=0.9\textwidth]{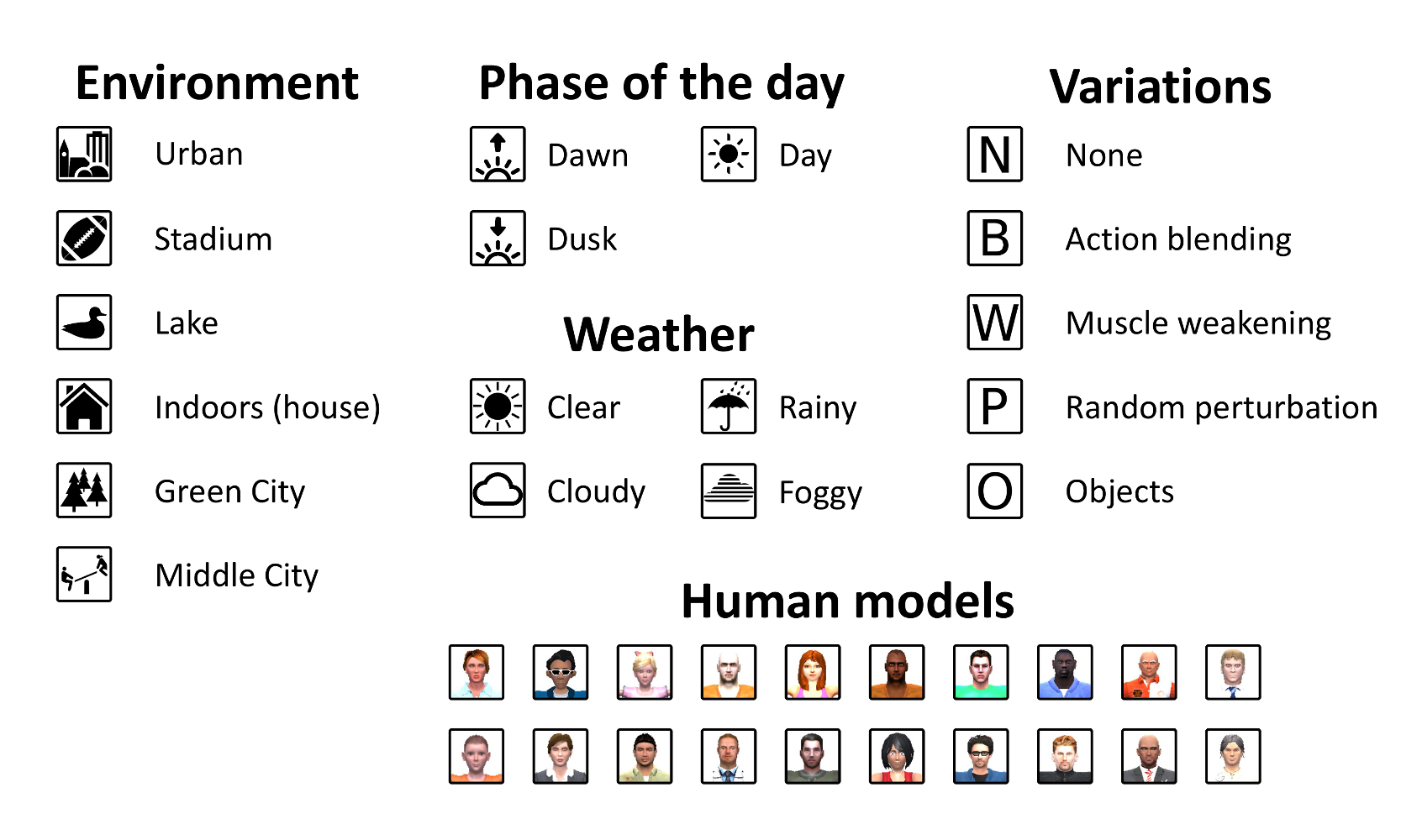}
    \caption{Legend for the variations shown in the video.}
    \label{fig:legend}
\end{figure*}

We have included a video (\cf Figure~\ref{fig:video}) as additional
supplementary material to our submission. The video shows random
subsamples for a subset of the action categories in \vhad. Each
subsample is divided into 5 main variation categories. 
Each video is marked with a label indicating the variation
being used, using the legend shown in Figure~\ref{fig:legend}.

\section{Conclusion}\label{sec:supp-conclusion}

Our detailed graphical model shows how a complex video generation 
can be driven through few, simple parameters. We have also
shown that generating action videos while still taking the effect
of physics into account is a challenging task. Nevertheless, we
have demonstrated that our approach is feasible through experimental
evidence on two real-world datasets, disclosing further information
about the performance of each RGB and optical flow channels in this
supplementary material.

\newpage\clearpage

\end{document}